\documentclass[letterpaper,11pt]{article}
\usepackage[margin=1in]{geometry}
\usepackage[bookmarks, colorlinks=true, plainpages = false, colorlinks=true,
citecolor=blue,
linkcolor=blue,
anchorcolor=red,
urlcolor=blue]{hyperref}
\usepackage{url}
\usepackage{amsmath,amssymb,amsfonts,amsthm}
\usepackage{algorithm}
\usepackage{algpseudocode}
\usepackage{dsfont}
\usepackage{color}
\usepackage{graphicx}
\usepackage{subfigure}
\usepackage{enumitem}
\usepackage{makecell}
\usepackage[font=small]{caption}

\DeclareMathOperator*{\minimize}{minimize}
\DeclareMathOperator*{\subj}{subject\,to}

\newtheorem{theorem}{Theorem}
\newtheorem{lemma}{Lemma}

\newtheorem{corollary}{Corollary}
\newtheorem{conjecture}{Conjecture}

\algnewcommand{\LeftComment}[1]{ \(\triangleright\) #1}

\newcommand{\Eb}{\mathbb{E}}
\newcommand{\Pb}{\mathbb{P}}

\allowdisplaybreaks

\usepackage[round]{natbib}

\newcommand{\mypara}[1]{{\smallskip \noindent \bf #1}\hspace{0.1in}}
\begin{document}

    \title{Federated Multi-armed Bandits with Personalization\footnotetext{To appear at the 24th International Conference on Artificial Intelligence and Statistics (AISTATS), 2021. }}
     \author{
     	Chengshuai Shi\thanks{Department of Electrical and Computer Engineering, University of Virginia, Charlottesville, VA, USA; Email:  \texttt{\{cs7ync,cong@virginia.edu\}}.}
    	 \qquad\qquad Cong Shen\footnotemark[1]
     	\qquad\qquad Jing Yang\thanks{School of Electrical Engineering and Computer Science, The Pennsylvania State University, University Park, PA, USA; Email: \texttt{yangjing@psu.edu}.} 
    }

    \date{\today}

    \maketitle
    
    \begin{abstract} 
     A general framework of personalized federated multi-armed bandits (PF-MAB) is proposed, which is a new bandit paradigm analogous to the federated learning (FL) framework in supervised learning and enjoys the features {of} FL with personalization. Under the PF-MAB framework, a mixed bandit learning problem that flexibly balances generalization and personalization is studied. A lower bound analysis for the mixed model is presented. We then propose the Personalized Federated Upper Confidence Bound (\textsc{PF-UCB}) algorithm, where the exploration length is chosen carefully to achieve the desired balance of learning the local model and supplying global information for the mixed learning objective. Theoretical analysis proves that \textsc{PF-UCB} achieves an $O(\log(T))$ regret regardless of the degree of personalization, and has a similar instance dependency as the lower bound. Experiments using both synthetic and real-world datasets corroborate the theoretical analysis and demonstrate the effectiveness of the proposed algorithm.
    \end{abstract}
      

\section{Introduction}
\label{sec:intro}

Federated learning (FL) is an emerging distributed machine learning paradigm that has many attractive properties. In particular, FL is motivated by the growing trend that massive amounts of real-world data are exogenously generated at edge devices, which are non-independent and identically distributed (non-IID) and highly imbalanced \citep{bonawitz2019towards}. FL focuses on many clients collaboratively training a machine learning model under the coordination of a central server while keeping the local data private at each client \citep{mcmahan2017communication}.

Earlier FL approaches focus on training a single global model that can perform well on the aggregated global dataset. However, the performance of the FL-trained global model on an individual client dataset degrades dramatically when significant heterogeneity among the local datasets exists, which raises the {concern} of using one global model for all individual clients in edge inference. To address this issue, FL with personalization \citep{smith2017federated} has been proposed. Instead of learning a single global model, each device aims at learning a mixture of the global model and its own local model \citep{hanzely2020federated,deng2020adaptive}, which provides an explicit trade-off between the two potentially competing learning goals.

While the main focus of the state of the art \emph{FL with personalization} is on the supervised learning setting, we propose to extend its core principles to the multi-armed bandits (MAB) problem. This is motivated by a corpus of practical applications, including:
\begin{itemize}[leftmargin=*]
\item  \textbf{Cognitive radio.} Consider a cellular network where one base station (BS) serves many devices (e.g., smartphones) that are geographically spread out in the coverage area. Each device wants to use the individually best channel (in terms of its own communication quality) for data transmission, while the BS wants to learn a globally best channel averaged over the coverage area (e.g., to broadcast control information). Since BS is fixed at one location, the global channel quality cannot be measured by the BS itself -- it has to come from measurements of the geographically distributed devices. However, if the devices make channel selection decisions only to learn the globally best channel, they may suffer from poor communication quality due to the local-global model mismatch. This calls for personalized federated MAB so that the global and local channel quality models are jointly considered.

\item \textbf{Recommender system.} Local servers want to recommend the most popular items to their served customers to maximize the expected rewards. The item popularity can only be learned via interacting with customers, leading to a bandit problem \citep{li2010contextual}. As different local servers have potentially heterogeneous customers, their local popularities are non-IID. In addition, each local server only collects data from a small group of customers, and the central server needs to average the locally learned popularity models to have a global model, without accessing the individual recommendation for privacy protection. However, the globally most popular item may not apply to the small group of customers of a particular local server, which again leads to personalization in a federated bandit setting, i.e., a joint consideration of global and local item popularities.
\end{itemize}

In both applications, the general FL characteristics need to be applied to an underlying bandit model, which greatly complicates the problem. The bandit setting is more difficult due to limited feedback (only observing one arm at a time) \citep{agarwal2020federated}. In addition, FL has a deterministic pipeline, while the data collection for bandit is {online} and {the} server-clients coordination becomes dynamic. {Moreover}, incorporating personalization represents another significant challenge, since a client has to consider other clients using the already limited bandit feedback. 

In this work, a novel framework of {personalized federated MAB (PF-MAB)} is developed, which represents a systematic attempt to bridge FL, MAB, and personalization. The PF-MAB framework generalizes the earlier works of federated bandits \citep{shi2021aaai,zhu2020federated,dubey2020differentially} and can serve as an umbrella for a variety of bandit problems that share the FL principles and are in need of personalization. In particular, we claim the following contributions.

\begin{itemize}[leftmargin=*]

\item A mixed global and local learning objective is studied in the PF-MAB framework, which allows for smoothly balancing \emph{generalization} and \emph{personalization} that depends on the specific application requirement.

\item We provide a lower bound analysis of the general PF-MAB model, which reveals the fundamental requirement of balancing global and local explorations.

\item Inspired by the lower bound analysis, we propose the {Personalized Federated Upper Confidence Bound} (\textsc{PF-UCB}) algorithm that carefully adjusts the lengths of local and global explorations based on the mixed learning objective. We also {address the synchronization problem caused by client heterogeneity by leveraging the exploration-exploitation tradeoff.} 

\item A rigorous regret upper bound analysis shows that \textsc{PF-UCB} achieves an $O(\log(T))$ regret \textit{regardless of the degree of personalization}, and has a similar instance dependency as shown in the lower bound. 

\item Additional algorithm enhancements guided by {the} theoretical analysis are also discussed. We verify the effectiveness and efficiency of \textsc{PF-UCB} via numerical experiments on both synthetic and real-world datasets. 

\end{itemize}

\section{Problem Formulation}
\label{sec:model}

\subsection{Single-player Stochastic MAB}

In the standard stochastic MAB setting, a single player directly plays $K$ arms, with rewards $X_k$ of arm $k\in[K]$ sampled independently from a $\sigma$-subgaussian distribution  with mean $\mu_{k}$. At time $t$, the player chooses an arm $\pi(t)$ and the goal is to maximize {the} expected cumulative reward in $T$ rounds, i.e., $\Eb\left[\sum_{t=1}^{T}X_{\pi(t)}(t)\right]$, which is characterized by minimizing the regret:
\begin{equation}
\label{eqn:regret_single}
R(T)=T\mu_*-\Eb\left[\sum\nolimits_{t=1}^{T}X_{\pi(t)}(t)\right],
\end{equation}
where $\mu_{*}:=\mu_{k_*}=\max\{\mu_1,...,\mu_K\}$. As shown by \cite{Lai:1985}, the regret is lower bounded by:
\begin{equation}\label{eqn:single_lower}
\liminf_{T\to\infty}\frac{R(T)}{\log(T)}\geq \sum\nolimits_{k\not=k_*}\frac{\mu_*-\mu_k}{\mathrm{kl}(X_k,X_{k_*})}
\end{equation}
where $\mathrm{kl}(X_k,X_{k_*})$ denotes the KL-divergence between the two corresponding distributions. 

\subsection{PF-MAB}
\mypara{Clients and local models.} In the PF-MAB framework, there are $M$ clients interacting with the same set of $K$ arms (referred as ``local arms''). The clients are labeled from $1$ to $M$ to facilitate the discussion (labelling is not used in the algorithm). For client $m$, arm $k$ generates local rewards $X_{k,m}(t)$ independently from a $\sigma$-subgaussian distribution with mean $\mu_{k,m}$. Without loss of generality, we assume $\sigma=1$. For different clients, their local models are non-IID, i.e.,  in general $\mu_{k,n}\neq \mu_{k,m}$ when $n \neq m$. A client can only interact with her own local MAB model by choosing arm $\pi_m(t)$ and receiving reward $X_{\pi_m(t),m}(t)$ at time $t$. Also, there is no direct communication between clients.

\mypara{The global model.} A global stochastic MAB model with the same set of $K$ arms (referred as ``global arms'') coexists with the local models, where the global reward $X_k(t)$ for the global arm $k$ is the average of local rewards, i.e., $X_k(t)=\frac{1}{M}\sum_{m=1}^MX_{k,m}(t)$. The global reward can be thought of as the \emph{virtual} {averaged} reward {had all $M$ clients pulled the same arm $k$ at time $t$}. Correspondingly, the mean reward of global arm $k$ is $\mu_k=\frac{1}{M}\sum_{m=1}^M\mu_{k,m}$. We note that although the global model is the average of local models, {the global rewards are not directly observable by any client}. 

\mypara{Communication.} In decentralized multi-player multi-armed bandits (MP-MAB), clients are prohibited to have \emph{explicit} communication with each other \citep{liu2010distributed,boursier2019sic}. We modify this constraint to enable client-server periodic communication that is similar to FL. Specifically, the clients can send ``local model updates'' to a central server, which then aggregates and broadcasts the updated ``global model'' to the clients. (We will specify these components later.) Note that just as in FL, communication is one of the major bottlenecks and the algorithm has to be conscious about its usage. This constraint is incorporated by imposing a loss $C$ each time a communication round happens, which will be accounted for in the regret. We also make the assumption that clients and server are fully synchronized \citep{mcmahan2017communication}.

\subsection{Personalization vs Generalization}
\subsubsection{Two Extreme Cases}
With the coexistence of local and global models, two extreme scenarios exist for the bandit learning: local-only and global-only. In the first case, clients only care about their own local performance, which is characterized by the local cumulative reward $r_l(T)$ as
\begin{equation*}
		r_l(T):=\Eb\left[\sum\nolimits_{t=1}^T\sum\nolimits_{m=1}^MX_{\pi_m(t),m}(t)\right].
\end{equation*} 
$r_l(T)$ is equivalent to the sum rewards of $M$ clients who play $M$ decoupled and non-interacting MAB games. Obviously, the optimal choice for client $m$ is arm $k_{*,m}$ with $\mu_{*,m}{:=\mu_{k_{*,m},m}} = \max_{k\in[K]}\mu_{k,m}$. However, only pursuing the locally optimal arm severely limits the ability of generalization across clients, especially when the degree of heterogeneity is significant. 

For the other extreme, clients only focus on learning the global model, which means to maximize the global cumulative reward: 
\begin{equation*}
	r_g(T) := \Eb\left[\sum\nolimits_{t=1}^T\sum\nolimits_{m=1}^MX_{\pi_m(t)}(t)\right].
\end{equation*}
In this case, although the client's action and observation are both on her local arms, the reward is defined with respect to the global arm \citep{shi2021aaai}. Ideally, the optimal choice to maximize $r_g(T)$ is to let all the clients play the optimal global arm $k_*$ with $\mu_*{:=\mu_{k_*}}=\max_{k\in[K]}{\mu_k}$. We note that this problem is recently proposed and studied in \cite{zhu2020federated,shi2021aaai}, which calls for efficient coordination among clients since no client can solve the global model individually. However, any efficient solution for this extreme case may lead to poor individual performance due to the non-IID local models. 

\subsubsection{Mixed Learning Objective}

To balance the need of both personalization and generalization, we hereby introduce a new learning objective which mixes $r_g(T)$ and $r_l(T)$ by a parameter $\alpha\in[0,1]$. This learning objective is referred to as the \emph{mixed cumulative reward}, which is defined as:
\begin{equation}\label{eqn:cumreward_personal}
	r(T):=\alpha r_l(T) +(1-\alpha) r_g(T).
\end{equation}
The parameter $\alpha$ provides a flexible choice of personalization: with $\alpha=1$, $r(T)$ becomes the sum rewards of $M$ individual single-player MAB games (full personalization); with $\alpha=0$, $r(T)$ only considers the global model (no personalization); with $0<\alpha<1$, both the global and local models are simultaneously taken into consideration by $r(T)$.

\subsubsection{Mixed Model}
An equivalent view of the mixed cumulative reward $r(T)$ in Eqn.~\eqref{eqn:cumreward_personal} is provided here, which facilitates our subsequent discussion. By unfolding $r_l(T)$ and $r_g(T)$, $r(T)$ can be rewritten as
\begin{equation*}
	r(T)  = \Eb\left[\sum\nolimits_{t=1}^T\sum\nolimits_{m=1}^MX'_{\pi_m(t),m}(t)\right],
\end{equation*}
where $X'_{\pi_m(t),m}(t)$ is a {hypothetical reward} that combines the local and global rewards, defined as:
\begin{equation}
	\label{eqn:mixed_reward}
		X'_{\pi_m(t),m}(t):=\alpha X_{\pi_m(t),m}(t)+(1-\alpha)X_{\pi_m(t)}(t).
\end{equation}	
Thus, maximizing the mixed cumulative reward can be equivalently viewed as playing a new MAB game with $X'_{k,m}(t)$ {as rewards} for the clients. However, since clients cannot directly observe the global reward, $X'_{k,m}(t)$ is only \emph{partially observable} at each individual client. We refer to this hypothetical game as the \textit{mixed model}. A similar reward definition using the weighted sum of clients' rewards has been adopted in \cite{branzei2019multiplayer}, albeit from a game theory perspective.

In client $m$'s mixed model, the mean reward $\mu_{k,m}':=\Eb\left[X'_{k,m}(t)\right]$ for arm $k$ can be calculated as:
\begin{equation}
	\label{eqn:eq_mean}
\mu'_{k,m}= \underbrace{\left(\alpha+\frac{1-\alpha}{M}\right)\mu_{k,m}}_{\text{local info}}+\underbrace{\frac{1-\alpha}{M}\sum\nolimits_{n\neq m}\mu_{k,n}}_{\text{global info}}.
\end{equation}
Since the global information in $\mu'_{k,m}$ is determined by other clients and cannot be accessed directly at client $m$, communication between clients and the server is of {critical} importance. 

With the mixed models, the notion of regret in Eqn.~\eqref{eqn:regret_single} can be generalized to $r(T)$ as
\begin{equation}\label{eqn:regret_personal}
	R(T) = T\sum_{m=1}^M\mu'_{*,m}-\Eb\left[\sum_{t=1}^T\sum_{m=1}^MX'_{\pi_m(t),m}(t)\right]+CMT_c,
\end{equation}
where $\mu'_{*,m}$ is the mean reward from the optimal arm $k'_{*,m}$ of client $m$'s mixed model, i.e., $\mu'_{*,m}:=\mu'_{k'_{*,m},m}=\max_{k\in[K]}\mu'_{k,m}$. The first term in $R(T)$ is the highest {expected} mixed cumulative reward that clients can get by always playing their optimal arms, which is similar to the optimal {expected} cumulative reward term $T\mu_*$ in Eqn.~\eqref{eqn:regret_single}. The additional loss term $CMT_c$ in Eqn.~\eqref{eqn:regret_personal} represents the communication loss, where $T_c$ is the total amount of communication slots. Without loss of generality, we assume that the optimal arm of each client on her mixed model is unique. {We further note that the optimal arms of different clients are likely to be different, because in general non-IID local models lead to $k'_{*,m}\neq k'_{*,n}$ when $m\neq n$.} We further denote $\Delta'_{k,m}=\mu'_{*,m}-\mu'_{k,m}$.

\section{Lower Bound Analysis}
\label{sec:lower}
A regret lower bound of PF-MAB is characterized in the following theorem.
\begin{theorem}\label{thm:fed_lower}
	For any consistent\footnote{{The consistent algorithm is defined the same way as in \cite{Lai:1985} but with the regret of Eqn.~\eqref{eqn:regret_personal}.}} algorithm $\Pi$, the regret $R(T)$ in Eqn.~\eqref{eqn:regret_personal} can be lower bounded as
	\begin{equation}
	\label{eqn:fed_lower}
		\liminf_{T\to\infty}\frac{R_{\Pi}(T)}{\log(T)}\geq \sum_{m=1}^M\sum_{k\neq k'_{*,m}}\max\bigg\{\frac{\Delta'_{k,m}}{\mathrm{kl}(Y_{k,m}, Y_{k'_{*,m},m})},  \frac{\Delta'_{k,m}}{\min\nolimits_{n: n\neq m,k'_{*,n}\neq k}\mathrm{kl}(Z^m_{k,n}, Z^m_{k'_{*,n},n})}\bigg\},
	\end{equation}
	where $Y_{k,m}=\left(\alpha+\frac{1-\alpha}{M}\right)X_{k,m}+\mu'_{k,m}-\left(\alpha+\frac{1-\alpha}{M}\right)\mu_{k,m}$ and $Z^m_{k,n}=\frac{1-\alpha}{M}X_{k,m}+\mu'_{k,n}-\frac{1-\alpha}{M}\mu_{k,m}$.
\end{theorem}
The proof of Theorem~\ref{thm:fed_lower} can be found in the appendix.  
{The communication cost is ignored in the analysis (i.e., $C=0$), but naturally, this lower bound still holds for $C>0$.} 
The lower bound in Eqn.~\eqref{eqn:fed_lower} sums over the maximum of two terms for all clients and suboptimal arms. First, random variable $Y_{k,m}$ with mean $\mu'_{k,m}$ represents an idealized degenerated game of client $m$'s mixed model where information from other clients, {i.e., $\{\mu_{k,l}\}_{ l\neq m}$}, is perfectly known. With $Y_{k,m}$, a lower bound for the regret of client $m$ learning arm $k$ for her mixed model can be obtained. Second, random variable $Z^m_{k,n}$ with mean $\mu'_{k,n}$ represents another idealized degenerated game of client $n$'s mixed model, where we assume full information of arm $k$ from all other clients except client $m$, {i.e., $\{\mu_{k,l}\}_{ l\neq m}$}. With $Z^m_{k,n}$, the regret of client $m$ providing information of arm $k$ to client $n$ is characterized. Then, building on this characterization, the regret of client $m$ providing information of arm $k$ to all other clients can be lower bounded by taking the worst case among them, i.e., the minimization term. This worst-case argument corresponds to the client who requires the most global information of arm $k$.  
To summarize, the first and second terms in the maximization characterize the necessary loss for learning local (for the client herself) and global (for all other clients) information of client $m$'s arm $k$, respectively. We also note that in the case of $\alpha=1$, i.e., local-only, Eqn.~\eqref{eqn:fed_lower} recovers the lower bound in Eqn.~\eqref{eqn:single_lower}, summed over $M$ local models.

More light can be shed on the lower bound by limiting the attention to Gaussian distributed rewards.
\begin{corollary}\label{col:fed_lower_gaussian}
	For any consistent algorithm $\Pi$, if the rewards follow Gaussian distributions with unit variance, the regret is lower bounded as
	\begin{equation*}
		\liminf_{T\to\infty}\frac{R_{\Pi}(T)}{\log(T)}\geq  \sum_{m=1}^M\sum_{k\neq k'_{*,m}}\max\bigg\{\frac{2\beta^2}{\Delta'_{k,m}}, \frac{2\gamma^2\Delta'_{k,m}}{(\Delta'_{k})^2}\bigg\},
	\end{equation*}
	where $\beta = \alpha+\frac{1-\alpha}{M}$, $\gamma = \frac{1-\alpha}{M}$ and $\Delta'_{k}=\min_{n:k'_{*,n}\neq k}\Delta'_{k,n}$.
\end{corollary}
Corollary \ref{col:fed_lower_gaussian} shows that the second term in the maximum is determined by $\Delta'_k$, which corroborates that the loss of learning global information for arm $k$ is determined by the hardest mixed model. 

We note that, as will be evident in the \textsc{PF-UCB} algorithm, the lower bound analysis reveals important guidelines for balancing global and local explorations. Nevertheless, neither Theorem \ref{thm:fed_lower} nor Corollary \ref{col:fed_lower_gaussian} establishes a \textit{universally tight} lower bound (for all $\alpha$). Characterizing the precise lower bound dependency on $\alpha$ is an interesting open problem, and we have the following conjecture.

\begin{conjecture}\label{con:fed_lower}
	For any consistent algorithm $\Pi$, as $T\to \infty$, $	\forall m\in[M]$ and  $\forall k: k\neq k'_{*,m}$, it holds that
	\begin{equation*}		
		\frac{\beta^2}{T_{k,m}}+\sum_{n: n\neq m, k'_{*,n}\neq k }\frac{\gamma^2}{T_{k,n}} \leq \frac{\eta^2\mathrm{kl}(X'_{k,m}, X'_{k'_{*,m},m})}{\log(T)},
	\end{equation*}
	where $T_{k,m}$ is the expected number of pulls on arm $k$ by client $m$ in the $T$ time slots, $\beta = \alpha+\frac{1-\alpha}{M}$, $\gamma = \frac{1-\alpha}{M}$ and $\eta = (\beta^2+(M-1)\gamma^2)^{\frac{1}{2}}$.
\end{conjecture}
Conjecture \ref{con:fed_lower} also recovers Eqn.~\eqref{eqn:single_lower} with $\alpha=1$. Furthermore, with $\alpha=0$ (global-only), it implies that $\liminf_{T\to\infty}\frac{R(T)}{\log(T)}\geq \sum_{k\neq k_{*}}\frac{M\Delta_k}{\mathrm{kl}(X_{k}, X_{k_{*}})}$, where $\Delta_k = \mu_*-\mu_k$. This result is reasonable as it is equivalent to the lower bound of a centralized client who directly maximizes the cumulative global reward.

\section{\textsc{PF-UCB} Algorithm}
\label{sec:alg}

\subsection{Challenges}

Solving the PF-MAB model faces several new challenges. The first challenge is that in order to maximize the mixed reward, both local and global information are essential. On one hand, the overall decision can be compromised (depending on the choice of $\alpha$) as long as one type of information is insufficiently learned. On the other hand, providing global information for other clients may degrade the individual performance since the additional exploration does not directly benefit the client. The key challenge is how to gain \emph{sufficient but not excessive} local and global information simultaneously based on the required degree of personalization.

A second challenge is that the game difficulties vary across clients. It is highly likely that different clients would need different amounts of global information. In other words, some clients may find their optimal arms much slower than the others, which is similar to the {client heterogeneity problem in FL \citep{li2020federated}. How to handle the resulting synchronization problem caused by client heterogeneity in PF-MAB becomes an important issue.}

Lastly, although communication is fundamental to providing global information, it incurs additional losses in regret. This benefit-cost balance needs to be addressed in the algorithm design. 

\subsection{Algorithm Design}

The Personalized Federated Upper Confidence Bound (\textsc{PF-UCB}) algorithm operates in phases (analogous to communication rounds in FL), and each phase consists of three sub-phases: global exploration, local exploration, and exploitation. {The set of  arms for global (resp. local) exploration are referred to as the set of global (resp. local) active arms. Specifically, at phase $p$, $A_m(p)$ (with cardinality $K_m(p)$) and $A(p)=\cup_{m\in[M]}A_m(p)$ (with cardinality $K(p)$) denote the set of local and global active arms respectively, which are both initialized as $[K]$.} \textsc{PF-UCB} for clients and the central server are presented in Algorithms \ref{alg:fedp_local} and \ref{alg:fedp_central}, respectively.

\begin{algorithm}
	\caption{\textsc{PF-UCB}: client $m$}
	\label{alg:fedp_local}
	\begin{algorithmic}[1]
		\Require $T$, $M$, $K$, $\alpha$
		\State Initialize $p\gets 1$; $A_m(1),A(1)\gets [K]$; $\forall k\in[K], s_{k,m}\gets 0$, $T_{k,m}\gets 0$; $g,h\gets 1$; {$O_m\gets 0$}
		\While{$A(p)\neq \emptyset$} 
		\Statex \LeftComment{\texttt{Global exploration}}
		\For{$g\leq K(p)\left\lceil(1-\alpha) f(p)\right\rceil$}
		\State $\pi\gets$ $(g\text{ mod }K(p))$-th arm in $A(p)$
		\State Pull arm $\pi$ and receive reward $r_\pi$
		\State $s_{\pi,m}\gets s_{\pi,m}+r_\pi$; $T_{\pi,m}\gets T_{\pi,m}+1$; $g\gets g+1$
		\EndFor
		\Statex \LeftComment{\texttt{Local exploration}}
		\For{$h\leq K_m(p)\lceil M\alpha f(p)\rceil$} 
		\State $\pi\gets$ $(h\text{ mod }K_m(p))$-th arm in $A_m(p)$
		\State Pull arm $\pi$ and receive reward $r_\pi$
		\State $s_{\pi,m}\gets s_{\pi,m}+r_\pi$; $T_{\pi,m}\gets T_{\pi,m}+1$; $h \gets h+1$
		\EndFor
		\State Update $\bar{\mu}_{k,m}(p)\gets s_{k,m}/T_{k,m}, \forall k\in A(p)$
		\State Send $\bar{\mu}_{k,m}(p), \forall k\in A(p) $ to the server
		\Statex \LeftComment{\texttt{Exploitation}}
		\If{$O_m = 0$}
		\State $\bar{k}'_{*,m}(p)\gets \arg\max_{k\in A_m(p)}\{\bar{\mu}'_{k,m}(p-1)\}$
		\Else \ $\bar{k}'_{*,m}(p)\gets O_m$
		\EndIf
		\State Pull arm $\bar{k}'_{*,m}(p)$ until receiving $\bar{\mu}_{k}(p), k\in A(p) $ 
		\State $\forall k\in {A_m(p)}, \bar{\mu}'_{k,m}(p)\gets \alpha\bar{\mu}_{k,m}(p)+(1-\alpha)\bar{\mu}_k(p)$
		\State Update $E_m(p)$ as in Eqn.~\eqref{eqn:ep} \Comment{\textit{Arm elimination}}
		\State $A_m(p+1)\gets A_m(p)\backslash E_m(p)$
		\If{$|A_m(p+1)|=1$} 
		\State {$O_m\gets$ the only arm in $A_m(p+1)$; $A_m(p+1)\gets\emptyset$}
		\EndIf
		\State Send $A_m(p+1)$ to the server
		\State Receive $A(p+1)$ from the server; $p\gets p+1$; $g,h\gets 1$
		\State {$K(p+1)\gets |A(p+1)|$, $K_m(p+1)\gets |A_m(p+1)|$}
		\EndWhile
		\State Stay on {arm $O_m$} until $T$ \Comment{\textit{Exploitation}}
	\end{algorithmic}
\end{algorithm}%
\begin{algorithm}[htb]
	
	\caption{\textsc{PF-UCB}: central server}
	\label{alg:fedp_central}
	\begin{algorithmic}[1]
		\Require $T$, $M$, $K$
		\State Initialize $p\gets 1$; $A(1)\gets [K]$
		\While{$A(p)\neq \emptyset$}
		\State Receive $\bar{\mu}_{k,m}(p), \forall k\in A(p)$ from all clients $m \in [M]$
		\State Update $\bar{\mu}_{k}(p)\gets \frac{1}{M}\sum_{m=1}^{M}\bar{\mu}_{k,m}(p)$, $\forall k\in A(p)$
		\State Send $\bar{\mu}_k(p),\forall k\in A(p)$ to all  clients
		\State Receive $A_m(p+1)$ from all  clients
		\State Send $A(p+1)\gets \cup_{m\in[M]}A_m(p+1)$ to all  clients
		\State  $p\gets p+1$
		\EndWhile
	\end{algorithmic}
\end{algorithm}

In phase $p$, global exploration is first performed in order to collect statistics to update the global information. Client $m$ explores each arm $k\in A(p)$, {i.e., global active arms,} for $n^g_{k,m}(p)=\lceil(1-\alpha)f(p)\rceil$ times, {and thus} the entire global exploration sub-phase lasts for $K(p)\lceil(1-\alpha)f(p)\rceil$ time slots. Note that $f(p)$ is a flexible exploration length {determined by the phase index $p$}, and its impact on the regret is analyzed later. Since all clients share the same global active arm set $A(p)$, the global exploration length is also the same for them. 

After the global exploration, the clients perform local exploration to update the local information. Each arm $k\in A_m(p)$ is played by client $m$ for $n^l_{k,m}(p)=\lceil M\alpha f(p)\rceil$ times, which means the local exploration lasts for $K_m(p)\lceil M\alpha f(p)\rceil$ time slots at client $m$. It is important to note that since different clients may have local exploration sets of different sizes, {i.e., $K_m(p)$ can be different across $m$}, the local exploration length may also vary across clients.

Note that the lengths of global and local explorations are carefully designed. For each arm $k\in A_m(p)$, it is explored for $\lceil(1-\alpha)f(p)\rceil$ times {during} global exploration (recall that $A_m(p)\subseteq A(p)=\cup_{m\in[M]}A_m(p)$) and $\lceil M\alpha f(p)\rceil$ times {during} local exploration, leading to {a total of} $n_{k,m}(p) = \lceil(1-\alpha)f(p)\rceil+\lceil M\alpha f(p)\rceil$ pulls by client $m$. At the same time, client $m$ is also assured that arm $k$ is pulled by every other client $n$ for at least $n^g_{k,n}(p)=\lceil(1-\alpha)f(p)\rceil$ times since they share the same $A(p)$. Thus, the proportion between local and global information is $\frac{(1-\alpha)+M\alpha }{(1-\alpha)}$, which coincides with the desired allocation in Eqn.~\eqref{eqn:eq_mean}\footnote{{A detailed discussion on matching $n_{k,m}(p) / n_{k,n}^g(p)$ with the weight ratio of Eqn.~\eqref{eqn:eq_mean} is provided in the appendix.}}.

After completing both global and local explorations, client $m$ first sends the updated local sample means {of all global active arms $k\in A(p)$, denoted as $\bar{\mu}_{k,m}(p)$ for arm $k$ at phase $p$,} as the ``local model updates'' to the server. Since the local exploration length may vary, the server may not receive the updates from all clients at the same time. Thus, it has to wait until the updated sample means {from all the clients} are received and then sends the aggregated ``global model'' $\bar{\mu}_k(p) = \frac{1}{M}\sum_{m=1}^M\bar{\mu}_{k,m}(p)$ back to the clients. While this waiting time is necessary to synchronize the clients, it also {leads to an increased regret}, i.e., all clients have to wait for the slowest client before the next iteration.

In PF-MAB, {The celebrated exploration-exploitation tradeoff in MAB is embraced to {keep the regret caused by this waiting time low}}. The idea is that clients who have already sent local updates can begin {\em exploitation} while the server still waits to collect information from other clients. Specifically, before $\bar{\mu}_k(p)$ are sent back, client $m$ keeps playing her empirically best arm  $\bar{k}'_{*,m}=\arg\max_{k\in A_m(p)}\{\bar{\mu}'_{k,m}(p-1)\}$, where $\bar{\mu}'_{k,m}(p-1)$ is the estimation of $\mu'_{k,m}$ in the preceding phase. Regret analysis shows that this is essential in keeping clients update periodically synchronized while achieving a low regret.

After the global sample means $\bar{\mu}_{k}(p)$ are broadcast to the clients, the estimation for $\mu'_{k,m}$ is updated as $\bar{\mu}'_{k,m}(p)=\alpha\bar{\mu}_{k,m}(p)+(1-\alpha)\bar{\mu}_{k}(p)$. Then, a local arm elimination procedure is performed such that the arms that are sub-optimal with a high probability are eliminated. With the newly calculated $\bar{\mu}'_{k,m}(p)$, the elimination set $E_m(p)$ can be constructed as:
\begin{equation}\label{eqn:ep}
    \left\{k: k\in A_m(p), \max_{l\in A_m(p)}\bar{\mu}'_{l,m}(p)-\bar{\mu}'_{k,m}(p)\geq 2B_p \right\},
\end{equation} 
where $B_p=\sqrt{{4\log(T)}/{(MF(p))}}$ is the confidence bound and $F(p)=\sum_{q=1}^p f(q)$. 
Note that the simple and clean form of $B_p$ comes from the carefully designed lengths of global and local explorations.  The local active set $A_m(p+1)$ for the next phase is updated as $A_m(p+1)=A_m(p)\backslash E_m(p)$. Finally, all the clients send $A_m(p+1)$ to the server and subsequently receive the global active set $A(p+1)=\cup_{m\in[M]}A_m(p+1)$ from the server. As long as an arm is in the local active set of at least one client, it is contained in the global active set because more global information regarding this arm is still needed to help (at least) that client make decisions. 

When the local active set contains only one arm, i.e., $|A_m(q)|=1$, client $m$ {marks the only left arm in $A_m(q)$ as the fixed arm $O_m$ and sets $A_m(q)=\emptyset$. Then, she only sends an empty set to the server for the local active set update
since her optimal arm is found.} Also, {with $A_m(q)=\emptyset$}, client $m$ does not perform local explorations any more. Nevertheless, global exploration is still necessary for client $m$ as long as $A(q)$ is not empty, because other clients still need information from her. {In the exploitation phase, she also directly plays the fixed arm $O_m$.} When all clients have found their optimal arms, i.e., $A(q)= \emptyset$, they all fixate on their identified arms until the end of $T$ without any further communication.

\textbf{Remarks.} It can be observed that the choice of local exploration length scales linearly with the number of clients, i.e., $n^l_{k,m}(p)=\lceil M\alpha f(p)\rceil\varpropto M$, which may not be desirable when $M$ is large. It is possible to simultaneously scale down the local and global exploration lengths as $M$ increases, e.g., $n^l_{k,m}(p)=\lceil\alpha f(p)\rceil$ and $n^g_{k,m}(p) = \left\lceil(1-\alpha)f(p)/M\right\rceil$, to further trade off exploration and communication, but this does not fundamentally change the regret behavior that is to be discussed. A final note is that only sample means and active sets are communicated in the entire procedure -- no raw samples and number of pulls are shared. This is similar to sharing model updates instead of raw data samples in FL, which helps preserve privacy. 

\section{Regret Analysis}
\label{sec:regret}
The theoretical analysis for \textsc{PF-UCB} is presented in this section. In particular, Theorem \ref{thm:regret_overall} characterizes a regret upper bound of \textsc{PF-UCB}.
\begin{theorem}\label{thm:regret_overall}
	$\forall m\in[M]$ and $\forall k\neq k'_{*,m}$, suppose $p'_{k,m}$ is the {smallest} integer that satisfies
	\begin{equation}
		\label{eqn:num_pull}
			MF(p'_{k,m})\geq \frac{64\log(T)}{(\Delta'_{k,m})^2}.
	\end{equation}
	The regret of \textsc{PF-UCB} can be bounded as
	{\begin{align}
		R(T) & \leq \sum\nolimits_{m=1}^M\sum\nolimits_{k\neq k'_{*,m}}\Delta'_{k,m}\sum\nolimits_{p=1}^{p'_{k,m}}\lceil \alpha M f(p) \rceil  +\sum\nolimits_{m=1}^M\sum\nolimits_{k\neq k'_{*,m}}\Delta'_{k,m}\sum\nolimits_{p=1}^{p'_{k}}\lceil (1-\alpha) f(p) \rceil\notag\\
		&+\sum\nolimits_{m=1}^M\sum\nolimits_{k\neq k'_{*,m}}\Delta'_{k,m}\sum\nolimits_{p=1}^{p'_{k,m}}K\left\lceil \alpha Mf(p)\right\rceil P'_{k,m}(p) + 2CMp'_{\max}+ 2(1+2C)M^2K, \label{eqn:regret_overall}
	\end{align}}
	where $p'_{k}=\max_{m\in[M]}\{p'_{k,m}\}$, $p'_{\max}=\max_{k\in[K]}\{p'_{k}\}$ and $P'_{k,m}(p)=\exp\{-\Delta^{'2}_{k,m}MF(p-1)/4\}$.
\end{theorem}
Detailed proof of Theorem~\ref{thm:regret_overall} can be found in the appendix, which shows that the total regret can be decomposed into local and global exploration losses, exploitation loss, and communication loss. {Note} that the local exploration loss (the first term) is determined individually by each client's local model, i.e., $p'_{k,m}$, while the global exploration loss (the second term) is determined globally, i.e., $p'_{k}$. This coincides with Theorem \ref{thm:fed_lower} and Corollary \ref{col:fed_lower_gaussian}. In addition, there is no global (resp. local) exploration loss in the local-only (resp. global-only) scenario, i.e., $\alpha=1$ (resp. $0$). Furthermore, although the constant term $2(1+2C)M^2K$ in Eqn.~\eqref{eqn:regret_overall} has a dependence on $M^2$, one may trade off this term with other regret terms by adjusting the confidence bound, e.g., specifying $B_{p} = \sqrt{4\log(MT)/(MF(p))}$.

{While Theorem \ref{thm:regret_overall} provides a general characterization with unspecified $f(p)$, the following corollary gives an explicit form of regret with $f(p)=2^p\log(T)$.}
\begin{corollary}\label{col:regret_overall_log}
	With $f(p)=2^p\log(T)$, it holds that
	\begin{equation*}
		R(T)= O\left(\sum_{m=1}^M\sum_{k\neq k'_{*,m}}\left[\frac{\alpha}{\Delta'_{k,m}}+\frac{\frac{1-\alpha}{M}\Delta'_{k,m}}{(\Delta'_{k})^2}\right]\log(T)\right),
	\end{equation*}
	where $\Delta'_{k}=\min_{n:k'_{*,n}\neq k}\{\Delta'_{k,n}\}$.
\end{corollary}
With this choice, \textsc{PF-UCB} achieves an $O(\log(T))$ regret \emph{regardless of $\alpha$}. It also has a similar instance dependency on $\Delta'_{k,m}$ and $\Delta'_{k}$ as shown in Corollary~\ref{col:fed_lower_gaussian}. Although the $\alpha$-dependency does not match Corollary \ref{col:fed_lower_gaussian}, which is not necessarily a tight lower bound, Corollary \ref{col:regret_overall_log} does match the sum of single-player lower bound when $\alpha=1$. {Interestingly, when $\alpha=0$, the achievable upper bound in Corollary \ref{col:regret_overall_log} approaches the conjectured lower bound in Conjecture \ref{con:fed_lower}}. It is also worth noting that the communication and exploitation losses when $f(p)=2^p\log(T)$ are both of order $O(1)$, which demonstrates its efficiency. Regrets with other choices of $f(p)$ can be found in the appendix.

We highlight the key components in the proof of Theorem \ref{thm:regret_overall} and Corollary~\ref{col:regret_overall_log} in the remainder of this section. A typical event 
\begin{equation*}
{G}=\{|\bar{\mu}'_{k,m}(p)-\mu'_{k,m}|\leq B_p, \forall p, \forall m\in[M], \forall k\in A_m(p)\}
\end{equation*}
is first established, and we can show that event ${G}$ happens with high probability.
\begin{lemma}\label{lem:tp_event}
	It holds that $\mathbb{P}({G}) := P_{G} \geq 1-\frac{2MK}{T}$.
\end{lemma}
We then analyze the different loss components of the total regret in the following.

\subsection{Exploration Loss}
First, Lemma~\ref{lem:num_pull} bounds the number of pulls at clients on their sub-optimal arms.
\begin{lemma}\label{lem:num_pull}
	Suppose event ${G}$ happens. For client $m$, sub-optimal arm $k\neq k'_{*,m}$ is guaranteed to be eliminated by phase $p'_{k,m}$ as defined in Theorem \ref{thm:regret_overall}.
\end{lemma}

Then, the local and global exploration losses, denoted as $R^{expr}_{l}(T)$ and $R^{expr}_{g}(T)$, respectively, can be bounded by the following lemma.
\begin{lemma}\label{lem:regret_expr}
	Suppose event ${G}$ happens. With $p'_{k,m}$ and $p'_{k}$ defined in Theorem \ref{thm:regret_overall}, $R^{expr}_{l}(T)$ and $R^{expr}_{g}(T)$ can be bounded, respectively, as
		\begin{equation*}
		\begin{aligned}
			&R^{expr}_{l}(T)\leq \sum_{m=1}^M\sum_{k\neq k'_{*,m}}\Delta'_{k,m}\sum_{p=1}^{p'_{k,m}}\lceil \alpha M f(p) \rceil,\\
			&R^{expr}_{g}(T)\leq \sum_{m=1}^M\sum_{k\neq k'_{*,m}}\Delta'_{k,m}\sum_{p=1}^{p'_{k}}\lceil (1-\alpha) f(p) \rceil.
		\end{aligned}
	\end{equation*}
\end{lemma}
Note that $R^{expr}_g(T)$ for arm $k$ is determined by $p'_k$, which is from the hardest local model for arm $k$. It also matches Theorem \ref{thm:fed_lower} and Corollary \ref{col:fed_lower_gaussian}.

\subsection{Exploitation Loss}
The exploitation loss $R^{expt}(T)$ is caused by the exploitations when a client has to wait for {other clients}. Noting that this loss stops once the optimal arm is declared. $R^{expt}(T)$ can be bounded as follows.
\begin{lemma}\label{lem:regret_expt}
	Suppose event ${G}$ happens. With $p'_{k,m}$ and  $P'_{k,m}(p)$ defined in Theorem \ref{thm:regret_overall}, $R^{expt}(T)$ can be bounded as
	\begin{equation*}
		R^{expt}(T)\leq \sum_{m=1}^M\sum_{k\neq k'_{*,m}}\Delta'_{k,m}\sum_{p=1}^{p'_{k,m}}K\left\lceil M\alpha f(p)\right\rceil P'_{k,m}(p).
	\end{equation*}
\end{lemma}

\subsection{Communication Loss}
Since communication stops once all the optimal arms are declared, the communication loss is bounded as:
\begin{lemma}\label{lem:regret_comm}
Suppose event ${G}$ happens. With $p'_{\max}$ defined in Theorem \ref{thm:regret_overall}, the communication loss $R^{comm}(T)$ can be bounded as $$R^{comm}(T)\leq 2CMp'_{\max}.$$
\end{lemma}

With Lemmas \ref{lem:tp_event} to \ref{lem:regret_comm}, Theorem \ref{thm:regret_overall} can be proved.

\section{Algorithm Enhancement}
\label{sec:alg_enhance}
While the exploration length in Section \ref{sec:alg} can be viewed as evenly splitting the workload among clients (especially for the global exploration), it ignores the fact that the same action results in different losses at different clients. We propose an enhancement to adaptively adjust the exploration lengths for client $m$, as follows:
\begin{equation*}
\begin{aligned}
	&n^l_{k,m}(p)\varpropto \frac{\alpha Mf(p)}{(\Delta'_{k,m})^{1/2}}, \forall k\in A_m(p), k\neq k'_{*,m};\\
	&n^g_{k,m}(p)\varpropto \frac{(1-\alpha)f(p)}{(\Delta'_{k,m})^{1/2}}, \forall k\in A(p), k\neq k'_{*,m}.
\end{aligned}
\end{equation*}
{More details on designing this enhancement can be found in the appendix.} 
Note that the exploration length for arm $k$ is now proportional to $1/(\Delta'_{k,m})^{1/2}$, which coincides with the intuition that the workload should decrease for those clients who suffer large losses, {i.e., with large $\Delta'_{k,m}$'s}. However, this is difficult to implement without the knowledge of $\Delta'_{k,m}$. One way to resolve this is to assume all of the sub-optimal gaps are the same, which results in the chosen length in Section \ref{sec:alg}. In this enhancement, however, we propose to replace $\Delta'_{k,m}$ by an estimation $\bar{\Delta}'_{k,m}(p)$ in phase $p$, which can be specified as $$\bar{\Delta}'_{k,m}(p)=\max_{l\in [K]}{\bar{\mu}'_{l,m}(p-1)}-\bar{\mu}'_{k,m}(p-1)+2B_{p-1}.$$ Rigorously analyzing the regret of this enhancement turns out to be difficult, and we evaluate it only through experiments.

\section{Experiments}
\label{sec:exp} 
Experiment results using both synthetic and real-world datasets are reported in this section to evaluate \textsc{PF-UCB} and the proposed  enhancement. The communication loss is set as $C=1$ and $f(p)$ is set to be $2^p\log(T)$. Details of the experiments (including the implementation codes) and additional results can be found in the appendix.

\begin{figure*}[htb]
    \centering
	\begin{minipage}[htb]{0.45\linewidth}
		\centering
		\includegraphics[width=0.9\linewidth]{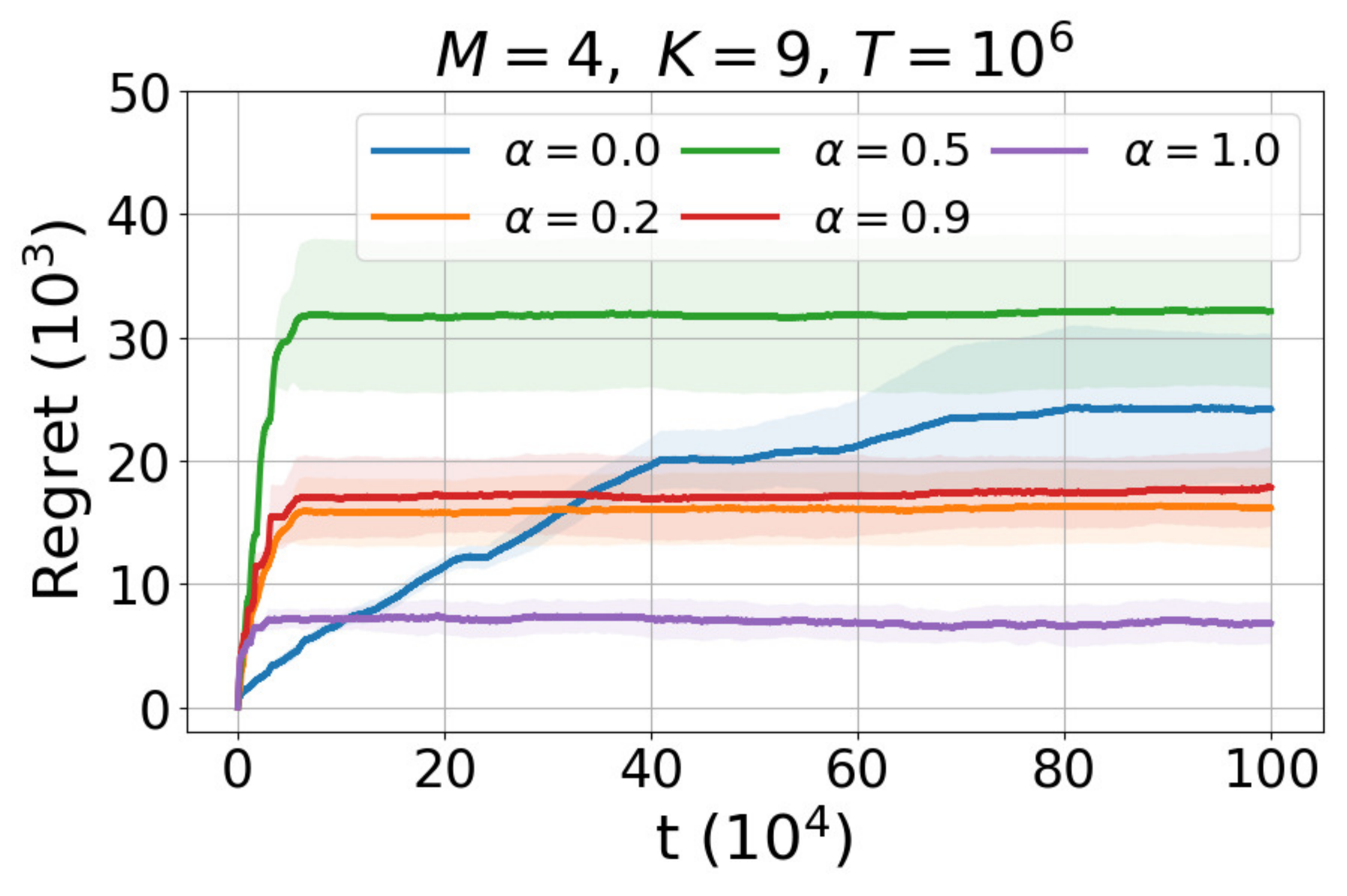}
		\caption{Synthetic Regret}
		\label{fig:syn_regret}
	\end{minipage}%
	\begin{minipage}[htb]{0.45\linewidth}
		\centering
		\includegraphics[width=0.9\linewidth]{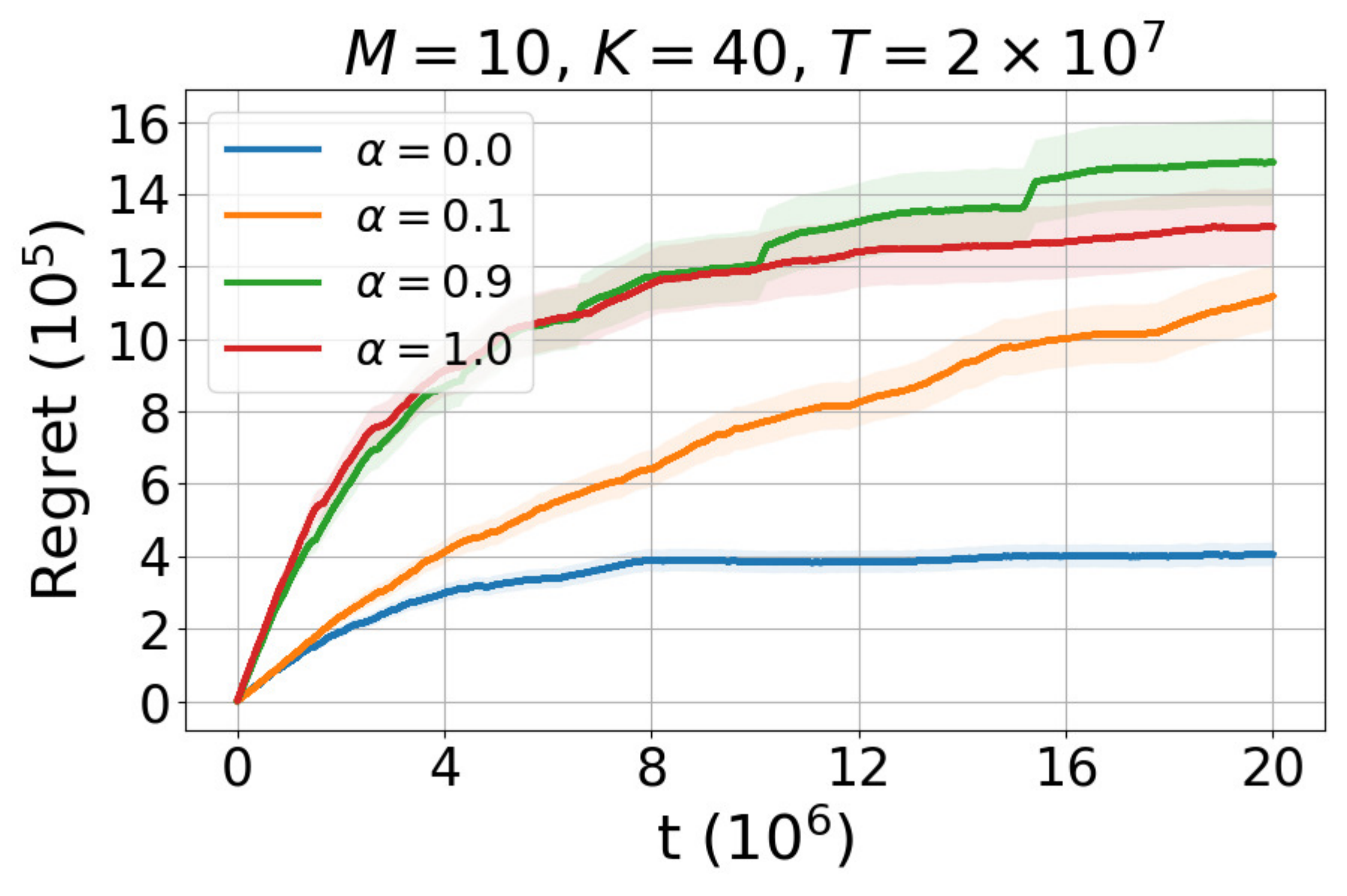}
		\caption{MovieLens Regret}
		\label{fig:movielens_regret}
	\end{minipage}
	\begin{minipage}[htb]{0.45\linewidth}
		\centering
		\includegraphics[width=0.9\linewidth]{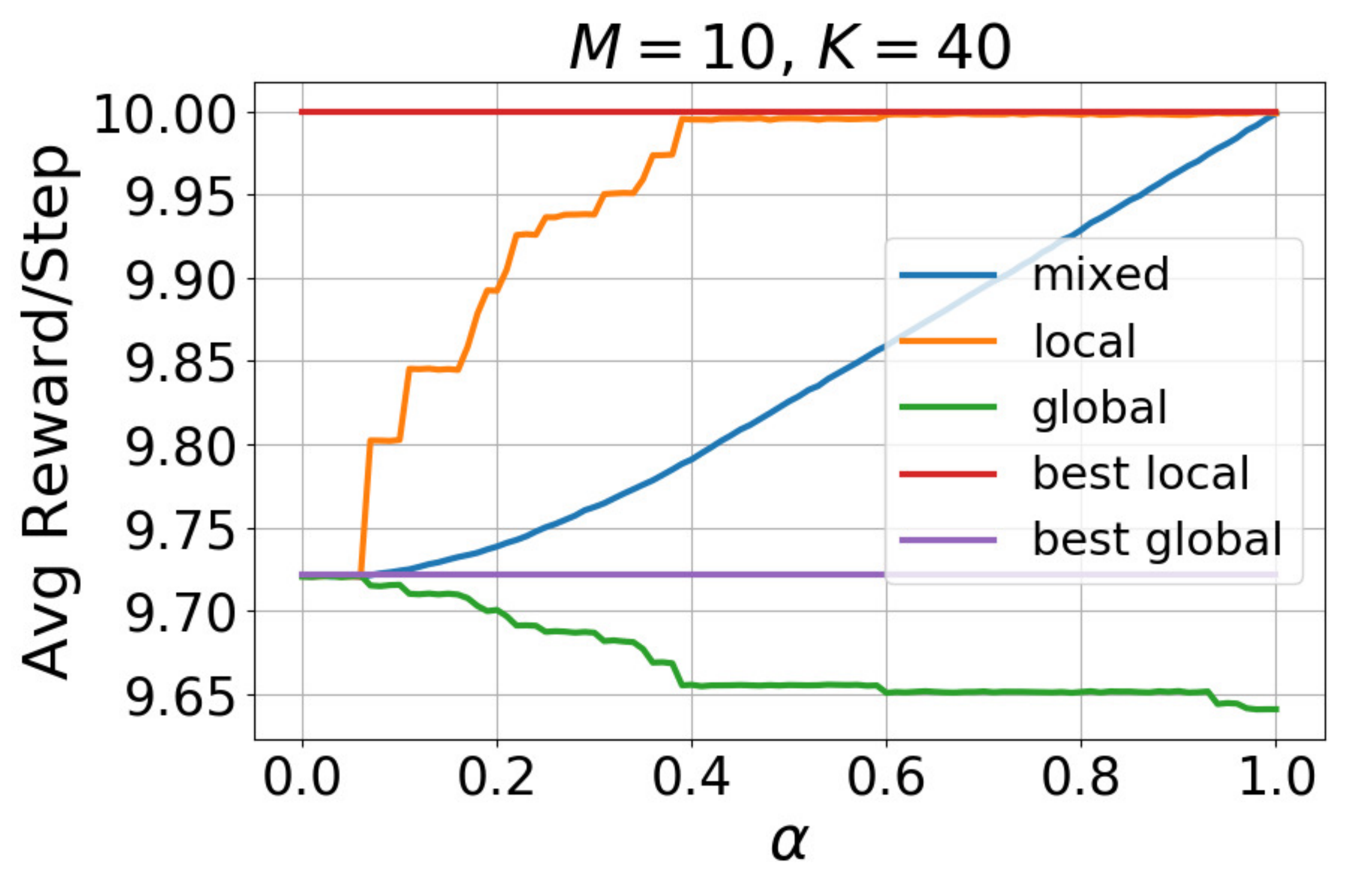}
		\caption{MovieLens Reward}
		\label{fig:movielens_reward}
	\end{minipage}
  	\begin{minipage}[htb]{0.45\linewidth}
		\centering
		\includegraphics[width=0.9\linewidth]{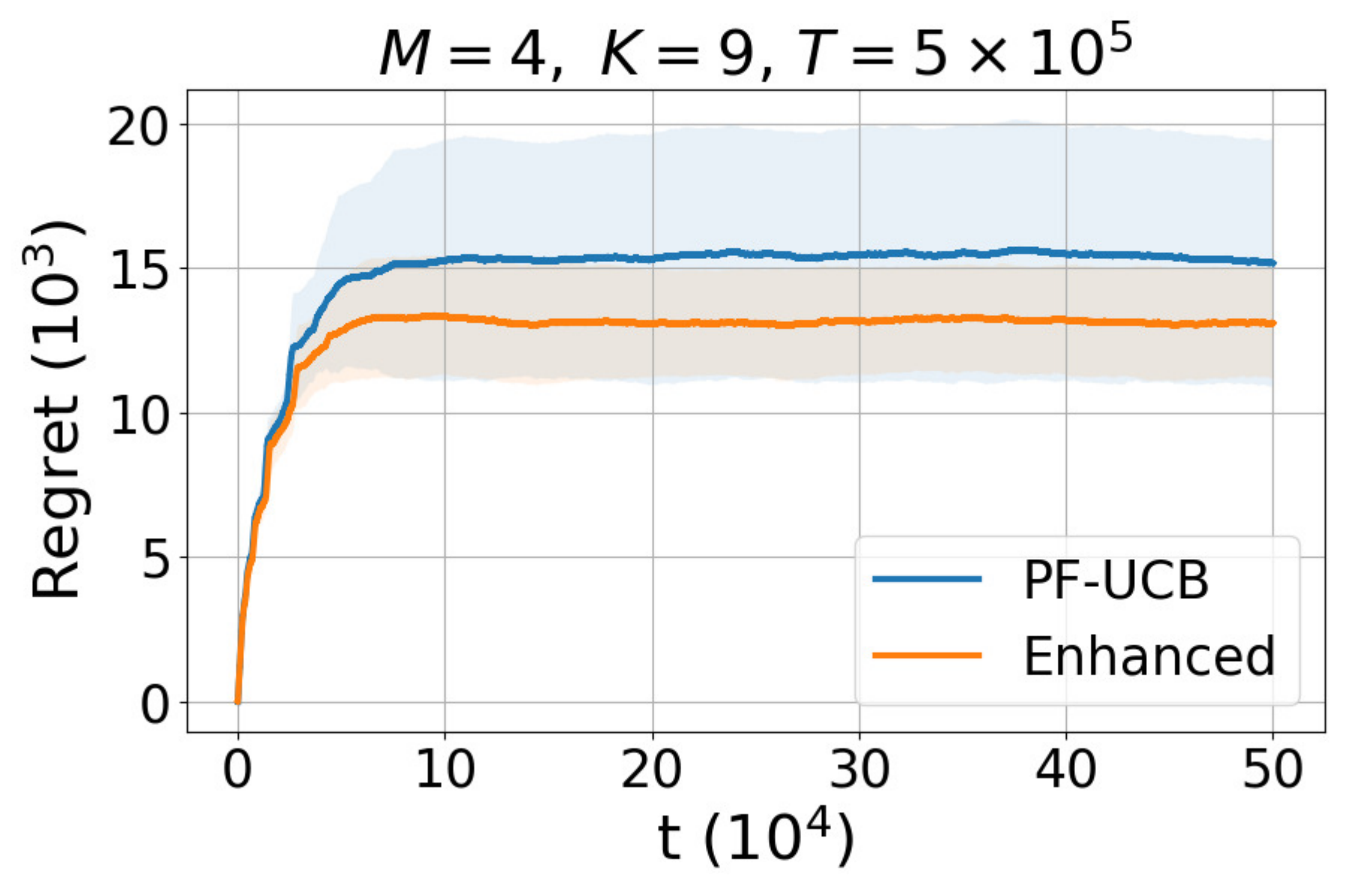}
		\caption{Enhancement}
		\label{fig:perf_enh}
	\end{minipage}%
\end{figure*}

First, \textsc{PF-UCB} is evaluated with various choices of $\alpha$ under a synthetic bandit game with $4$ clients and $9$ arms. The game is carefully designed such that all clients have different local optimal arms and the global optimal arm is also sub-optimal {locally}. Fig.~\ref{fig:syn_regret} shows that \textsc{PF-UCB} successfully converges to the optimal choices across different values of $\alpha$, which proves its effectiveness in handling different combinations of personalization and generalization. The varying overall regrets and convergence speeds are the result of different game difficulties associated with different $\alpha$. 

We then return to one of the motivating examples -- the recommender system -- and utilize the real-world MovieLens dataset \citep{Cantador:RecSys2011} for an empirical study of PF-MAB. {The $2113$ clients and $10197$ movies in the dataset are randomly divided into $10$ and $40$ groups, respectively, and the averaged movie ratings from each group of clients are used to construct their local rewards, which vary across the groups of clients and naturally lead to non-IID local models.} This game is larger and  harder than the previous synthetic game. Especially, some groups have suboptimality gaps on their mixed models at around $10^{-4}$. As shown in Fig.~\ref{fig:movielens_regret}, sub-linear regrets are achieved by \textsc{PF-UCB} with different values of $\alpha$. Note that in some cases (e.g., $\alpha=0.1$), the algorithm does not completely converge within the given horizon; however, the regret curve only increases slowly at the end, which suggests that most of the suboptimal arms are eliminated.\footnote{We note that in practice, the dataset is likely to be structured more carefully, e.g., grouping movies by categories instead of randomly, which would in general lead to easier games and faster convergence.}

We also evaluate the rewards instead of regrets in the same MovieLens experiments. Fig.~\ref{fig:movielens_reward} reports the averaged per-step reward that \textsc{PF-UCB} achieves with varying $\alpha$. The optimal global and local rewards (labeled as ``best global'' and ``best local'') represent the theoretically highest {global and local} mean rewards, respectively. {The mixed, global, and local rewards (labeled as ``mixed'', ``global'', and ``local'')} generated by the actions of clients with \textsc{PF-UCB} are plotted. Fig.~\ref{fig:movielens_reward} shows that the mixed and global rewards almost meet the optimal global rewards with $\alpha=0$ (global-only), while the local rewards are highly sub-optimal. With an increase of $\alpha$, the mixed and local rewards trend up, indicating the focus is gradually shifted towards the local rewards, and simultaneously the global rewards trend down. At $\alpha=1$ (local-only), the mixed and local rewards almost achieve the optimal local rewards, while the global rewards are poor. This gradual shifting shows that introducing  $\alpha$  provides a smooth tradeoff between local and global rewards.

Lastly, the algorithm enhancement in Section \ref{sec:alg_enhance} is evaluated. With a $4$-client $9$-arm game, the performance of the original and enhanced \textsc{PF-UCB} is compared in Fig.~\ref{fig:perf_enh} with $\alpha=0.5$. It can be observed that both algorithms converge but the enhanced design has a lower regret, demonstrating its effectiveness.

\section{Discussions}

The proposed PF-MAB framework, the PF-UCB algorithm, and their companion theoretical analysis represent our initial attempt to bridge FL, MAB, and personalization. It also leaves several interesting directions for future research.

First, PF-UCB falls in the category of ``cross-silo'' federated learning \citep{kairouz2019advances}, where clients are fixed, reliable and always available. Another important category of FL is the ``cross-device'' model, where only a fraction of clients are available at one time and some of them may fail or drop out. It is interesting to study the PF-MAB framework in such cross-device setting. Some simple modifications can be made to PF-UCB for this setting, e.g., sampling from the randomly available clients as participants. It is however challenging to have a rigorous theoretical analysis with time-varying participating clients.

Second, as stated in Section~\ref{sec:lower}, it would be of great value to have a \emph{tight} lower bound analysis under the PF-MAB model, and (dis)proving Conjecture~\ref{con:fed_lower} may serve as a starting point. We believe the main difficulty comes from that, instead of the traditional exploration-exploitation tradeoff, learning in PF-MAB actually faces a much more complicated tradeoff among global exploration, local exploration, and exploitation. Moreover, communication loss is ignored in Theorem~\ref{thm:fed_lower} and Conjecture~\ref{con:fed_lower}, and it would further complicate the tradeoff if communication loss is considered.

Last but not the least, as the algorithm enhancements proposed in Section~\ref{sec:alg_enhance} are validated only through experiment, it would be valuable to have rigorous analysis on whether (and how much) it outperforms the original PF-UCB.

\section{Related Works}
\label{sec:related}

\mypara{Differences to FL.} FL has been an active research area over the past few years \citep{mcmahan2017communication,bonawitz2019towards}, with many attractive features as discussed in Section~\ref{sec:intro}. See \cite{kairouz2019advances,li2020federated} for comprehensive surveys of FL. In particular, FL with personalization \citep{kulkarni2020survey} is an emerging topic, where learning a mixed local and global model (as in PF-MAB) is a representative approach \citep{hanzely2020federated,deng2020adaptive,mansour2020three} among others \citep{wang2019federated, smith2017federated, jiang2019improving,fallah2020personalized}. Nevertheless, existing studies on FL are almost exclusively on supervised learning and there is very limited literature considering bandit \citep{shi2021aaai,zhu2020federated}. 

\mypara{Differences to {Multi-player} MAB.} 
The decentralized MP-MAB problem is related to PF-MAB but fundamentally different. The MP-MAB research considers either the ``cooperative''  setting \citep{landgren2016distributed,landgren2018social,wang2019distributed} or the ``competitive'' setting \citep{rosenski2016multi,boursier2019sic,Shi2020aistats}. Although user-dependent local models are studied in both settings \citep{shahrampour2017multi, bistritz2018distributed,boursier2019practical}, our work is the first to study a flexible and mixed learning objective with partially observable rewards in MAB, to the best of our knowledge.

\mypara{Recent Advances.} A few recent works have touched upon the concept of federated bandits but none of them systemically addresses the key challenges {introduced by personalization}. \cite{li2020federated2,dubey2020differentially} assume IID local models and focus on privacy protection. \cite{agarwal2020federated} studies regression-based contextual bandits as an example of the federated residual learning framework, which does not generalize to our formulation. The recent studies in \cite{zhu2020federated, shi2021aaai} are more related to this work, where federated MAB without personalization (i.e., global-only) is studied. A similar client-server communication protocol is adopted in \cite{shi2021aaai} while a gossiping information-sharing strategy is applied in \cite{zhu2020federated}.

\section{Conclusions}
\label{sec:con}
In this work, we have developed a general PF-MAB framework to bridge MAB, FL, and personalization. By focusing on learning a mixed global-local objective, this framework enables a flexible tradeoff between personalization and generalization. A lower bound analysis was provided for PF-MAB. The proposed \textsc{PF-UCB} algorithm caters to the need of personalization and addresses {client heterogeneity} by leveraging the exploration-exploitation tradeoff. Theoretical analysis showed that \textsc{PF-UCB} can achieve a regret of $O(\log(T))$ regardless of the degree of personalization, and share a similar instance dependency as the lower bound. Numerical experiments on both synthetic and real-world datasets proved the effectiveness of the proposed algorithms and corroborated the theoretical analysis. 
PF-MAB is a new bandit framework that introduces the fundamental twist between local and global learning on top of the classical exploration-exploitation tradeoff, and sets the stage for potential future research activities.

\newpage
\appendix

\section{Details of Choosing Exploration Lengths and Algorithm Enhancement}\label{supp:enhance}
As stated in Section \ref{sec:alg}, the key challenge to solve PF-MAB is how to gain \emph{sufficient but not excessive} local and global information simultaneously based on the required degree of personalization. Sections \ref{sec:alg} and \ref{sec:alg_enhance} provide two choices and here the details behind these choices are elaborated.

From client $m$'s perspective on a locally active arm $k\neq k'_{*,m}$, in order to maintain the convergence rate of ${1}/{(MF(p))}$ (as specified in Section \ref{sec:alg}) while reducing the loss, an optimization problem over $N_{k,m}(p)$ and $ N^g_{k,n}(p), \forall n\neq m$ can be formulated as:
\begin{align*}
	&\minimize{\,\, N_{k,m}(p)\Delta'_{k,m}+\sum\nolimits_{n\neq m, k'_{*,n}\neq k} N^g_{k,n}(p)\Delta'_{k,n}} \\
	&\subj{\,\, \frac{\left[\alpha+(1-\alpha)/M\right]^2}{N_{k,m}(p)}+\sum_{n\neq m}\frac{\left[(1-\alpha)/M\right]^2}{N^g_{k,n}(p)}\leq \frac{1}{MF(p)}}
\end{align*}
where $N_{k,m}(p)$ is the number of pulls on arm $k$ at client $m$ up to phase $p$, and $N^g_{k,n}(p)$ is the guaranteed number of global pulls on arm $k$ at a different client $n$ up to phase $p$. The optimization objective is the loss associated with client $m$'s local and global information estimation for arm $k$, while the constraint is a sufficient condition for $B_p=\sqrt{{4\log(T)}/{(MF(p))}}$ and Lemma \ref{lem:tp_event} to hold. Note that the convergence rate constraint can have many forms, and the choice here is to match the discussion in the main paper.

Using the Cauchy-Schwarz inequality, the exploration length described in Section \ref{sec:alg_enhance} can be obtained as:
\begin{equation*}
\begin{aligned}
	&n^l_{k,m}(p)\varpropto \frac{\alpha Mf(p)}{(\Delta'_{k,m})^{1/2}}, \forall k\in A_m(p), k\neq k'_{*,m};\\
	&n^g_{k,m}(p)\varpropto \frac{(1-\alpha)f(p)}{(\Delta'_{k,m})^{1/2}}, \forall k\in A(p), k\neq k'_{*,m}, 
\end{aligned}
\end{equation*}
and $N^l_{k,m}(p)=\sum_{q=1}^p n^l_{k,m}(q)$, $N^g_{k,m}(p)=\sum_{q=1}^p n^g_{k,m}(q)$ and $N_{k,m}(p)=N^l_{k,m}(p)+N^g_{k,m}(p)$. This result is the key to choosing exploration lengths as it builds up the relationship between local and global explorations.

The issue however is that the knowledge of $\Delta'_{k,m}$ is unavailable. An easy way to tackle this problem is to assume all the sub-optimal gaps are the same, which results in the chosen length in \textsc{PF-UCB} in Section \ref{sec:alg}. The alternative way proposed in Section \ref{sec:alg_enhance} is to use $\bar{\Delta}'_{k,m}(p)=\max_{l\in [K]}{\bar{\mu}'_{l,m}(p-1)}-\bar{\mu}'_{k,m}(p-1)+2B_{p-1}$ in place of $\Delta'_{k,m}(p)$. This approach leverages information collected in the game. However, $\bar{\Delta}'_{k,m}(p)$ needs to be communicated to the server and then broadcast to maintain synchronization among clients, which may increase the risk of privacy leaking.

\section{Proof for the Lower Bound Analysis in Theorem \ref{thm:fed_lower}}
\begin{proof}
	First, the following lemma recalls the classic result from the single-player MAB \citep{Lai:1985}, which directly leads to the lower bound in Eqn.~\eqref{eqn:single_lower}.
	\begin{lemma}\label{lem:pull_single}
		For any consistent policy $\Pi$, for any arm $k$ such that $\mu_{k}<\mu_{k_*}$, it holds that
		\begin{equation*}
			\liminf_{T\to\infty} \frac{T_{k}}{\log(T)}\geq \frac{1}{\mathrm{kl}\left(X_k,X_{k_*}\right)},
		\end{equation*}
		where $T_{k}$ is the expected number of pulls performed on arm $k$ during $T$.
	\end{lemma}
	Then, from client $m$'s perspective of her suboptimal arm $k\neq k_{*,m}$ on the mixed model, the mixed reward in Eqn.~\eqref{eqn:mixed_reward} can be decomposed as
	\begin{equation*}
		X'_{k,m} = \left(\alpha+\frac{1-\alpha}{M}\right)X_{k,m}+\frac{1-\alpha}{M}\sum_{n\neq m}X_{k,n}.
	\end{equation*}
	The difficulty is that $X'_{k,m}$ involves the rewards from all $M$ clients, which are $M$ sources of randomness. Next we attempt to isolate these sources of randomness.
	
	First, if we assume client $m$ has perfect knowledge of $\{\mu_{k,n}\}_{n\neq m}$, a new random variable $Y_{k,m}$ is constructed as
	\begin{equation*}
		Y_{k,m}=\left(\alpha+\frac{1-\alpha}{M}\right)X_{k,m}+\frac{1-\alpha}{M}\sum_{n\neq m}\mu_{k,n} = \left(\alpha+\frac{1-\alpha}{M}\right)X_{k,m}+\mu'_{k,m}-\left(\alpha+\frac{1-\alpha}{M}\right)\mu_{k,m}.
	\end{equation*}
	Under this construction, $Y_{k,m}$ shares the same mean with $X'_{k,m}$ while the randomness only comes from $X_{k,m}$. Then, $Y_{k,m}$ forms a new hypothetical bandit game degenerated from client $m$'s mixed model, where the mean rewards and the optimal arm remain the same. With Lemma \ref{lem:pull_single}, if client $m$ individually interacts with this new game, her pulls on arm $k$ can be bounded as
	\begin{equation*}
		\liminf_{T\to\infty} \frac{T_{k,m}}{\log(T)}\geq \frac{1}{\mathrm{kl}\left(Y_{k,m},Y_{k'_{*,m},m}\right)}.
	\end{equation*}

	On the other hand, from a different client $n$'s perspective, whose arm $k$ is also sub-optimal, she also needs information of client $m$'s arm $k$. However, client $n$'s mixed reward is constructed as
	\begin{equation*}
		X'_{k,n} = \left(\alpha+\frac{1-\alpha}{M}\right)X_{k,n}+\frac{1-\alpha}{M}X_{k,m}+\frac{1-\alpha}{M}\sum_{l\neq m,n}X_{k,l},
	\end{equation*}
	which is different from $X'_{k,m}$. Following a similar idea of isolating randomness, if we assume client $n$ has perfect knowledge of $l\neq m,\mu_{k,l}$, including $\mu_{k,n}$, a new random variable $Z^m_{k,n}$ can be constructed as
	\begin{equation*}
		Z^m_{k,n} = \left(\alpha+\frac{1-\alpha}{M}\right)\mu_{k,n}+\frac{1-\alpha}{M}X_{k,m}+\frac{1-\alpha}{M}\sum_{l\neq m,n}\mu_{k,l}=\frac{1-\alpha}{M}X_{k,m}+\mu'_{k,n}-\frac{1-\alpha}{M}\mu_{k,m}.
	\end{equation*}
	Under this construction, $Z^m_{k,n}$ shares the same mean as $X_{k,n}$ while the randomness only comes from $X_{k,m}$. Then $Z^m_{k,n}$ forms another new hypothetical bandit game degenerated from client $n$'s mixed model, where the optimal arm remains the same and client $m$ has to provide information to help client $n$ distinguish arm $k$. Similarly, with Lemma \ref{lem:pull_single}, if client $m$ individually interacts with this new game, her pulls on arm $k$ can be bounded as
	\begin{equation*}
		\liminf_{T\to\infty} \frac{T_{k,m}}{\log(T)}\geq \frac{1}{\mathrm{kl}\left(Z^m_{k,n},Z^m_{k'_{*,n},n}\right)}.
	\end{equation*}
	
	Since $Z^m_{k,n}$ can be constructed for any client, it must hold that
	\begin{equation*}
		\liminf_{T\to\infty} \frac{T_{k,m}}{\log(T)}\geq \max_{n:n\neq m, k'_{*,n}\neq k}\left\{ \frac{1}{\mathrm{kl}\left(Z^m_{k,n},Z^m_{k'_{*,n},n}\right)}\right\}  =  \frac{1}{\min_{n:n\neq m, k'_{*,n}\neq k}\left\{\mathrm{kl}\left(Z^m_{k,n},Z^m_{k'_{*,n},n}\right)\right\}}.
	\end{equation*}

	Combining the above results, we can have
	\begin{equation*}
		\liminf_{T\to\infty} \frac{T_{k,m}}{\log(T)}\geq \max\left\{\frac{1}{\mathrm{kl}\left(Y_{k,m},Y_{k'_{*,m},m}\right)}, \frac{1}{\min_{n:n\neq m, k'_{*,n}\neq k}\left\{\mathrm{kl}\left(Z^m_{k,n},Z^m_{k'_{*,n},n}\right)\right\}}\right\}.
	\end{equation*}
	Since the regret can be decomposed as
	\begin{equation*}
		R(T) = \sum_{m=1}^M\sum_{k: k\neq k'_{*,m}}T_{k,m}\Delta'_{k,m},
	\end{equation*}
	Theorem \ref{thm:fed_lower} can be established.
\end{proof}
Note that the randomness isolation utilized in the proof reduces the hardness of the problem, which results in a relaxed lower bound. Although it can recover the single-player stochastic MAB lower bound with $\alpha=1$, when $\alpha$ moves away from $1$, the lower bound becomes less tight. 

\section{Discussions for Theorem \ref{thm:regret_overall}}
\begin{table}[htb]
	\centering
	\begin{tabular}{lll}
		\hline
		$f(p)$        & $p_{k,m}$, $k\not=k'_{*,m}$  & $R(T)$\\
		\hline
		\medskip
		$\lambda$   & $O\left(\frac{\log(T)}{M\lambda(\Delta'_{k,m})^2}\right)$ & $O\left(\sum_{m=1}^M\sum_{k\neq k'_{*,m}} \left[\frac{\alpha}{\Delta'_{k,m}}+ \frac{\frac{1-\alpha}{M}\Delta'_{k,m}}{\Delta^{'2}_{k}}\right]\log(T)+\frac{C\log(T)}{\lambda(\Delta'_{\min})^2}\right)$   \\
		\medskip
		$\lambda\log(T)$   & $O\left(\frac{1}{M\lambda(\Delta'_{k,m})^2}\right)$ & $O\left(\sum_{m=1}^M\sum_{k\neq k'_{*,m}} \left[\frac{\alpha}{\Delta'_{k,m}}+ \frac{\frac{1-\alpha}{M}\Delta'_{k,m}}{\Delta^{'2}_{k}}\right]\log(T)+\frac{C}{\lambda(\Delta'_{\min})^2}\right)$   \\
		\medskip
		$2^{p}$   & $O\left(\log\left(\frac{\log(T)}{M(\Delta'_{k,m})^2}\right)\right)$ & $O\left(\sum_{m=1}^M\sum_{k\neq k'_{*,m}} \left[\frac{\alpha}{\Delta'_{k,m}}+ \frac{\frac{1-\alpha}{M}\Delta'_{k,m}}{\Delta^{'2}_{k}}\right]\log(T)+CM\log\left(\frac{\log(T)}{M(\Delta'_{\min})^2}\right)\right)$   \\
		\medskip
		$2^{p}\log(T)$   & $O\left(\log\left(\frac{1}{M(\Delta'_{k,m})^2}\right)\right)$ & $O\left(\sum_{m=1}^M\sum\nolimits_{k\neq k'_{*,m}}\left[\frac{\alpha}{\Delta'_{k,m}}+\frac{\frac{1-\alpha}{M}\Delta'_{k,m}}{(\Delta'_{k})^2}\right]\log(T)+CM\log\left(\frac{1}{M(\Delta'_{\min})^2}\right)\right)$   \\
		\hline
	\end{tabular}\\
	\caption{Regret of \textsc{PF-UCB} algorithm with different choices of $f(p)$}
	\label{tbl:regret_P_Fed_UCB}
	{$\lambda$ is a constant; $\Delta'_k=\min_{n: k'_{*,n}\neq k}\{\Delta'_{k,n}\}$; $\Delta'_{\min}=\min_{k}\{\Delta'_{k}\}$.}
\end{table}
Table \ref{tbl:regret_P_Fed_UCB} summarizes the regrets under several different choices of $f(p)$,  including $f(p)=2^p\log(T)$ in Corollary \ref{col:regret_overall_log}. All choices listed in Table \ref{tbl:regret_P_Fed_UCB} achieve a similar exploration regret and a non-dominating exploitation loss (which is omitted in the regret expression). However, they lead to varying communication losses. With $f(p)=\lambda$, the communication loss is of order $O(\log(T))$ and scales with ${1}/{(\Delta'_{\min})^2}$, which actually dominates the exploration loss. This is the result of the unnecessary communications with $f(p)=\lambda$. With $f(p)=\lambda\log(T)$, the communication loss is no longer of order $O(\log(T))$; however, it still scales with ${1}/{(\Delta'_{\min})^2}$. The dependency of communication loss on $\Delta'_{\min}$ is improved with an exponential $f(p)$, as both $f(p)=2^p$ and $f(p)=2^p\log(T)$ have communication losses that scale only with $\log\left({1}/{\Delta'_{\min}}\right)$, which greatly reduces the communication burden. Furthermore, with $f(p)=2^p\log(T)$, the communication cost is a constant that is independent of $T$. Thus, among all considered choices of $f(p)$, the most preferable one is $f(p)=2^p\log(T)$.

{We further note that all the choices of $f(p)$ listed in Table~\ref{tbl:regret_P_Fed_UCB} do not depend on the communication loss parameter $C$. This is made to simplify the problem, as otherwise the analysis will have a convoluted relationship between the exploration loss and the communication loss. Intuitively, with a larger $C$, it is better to increase $f(p)$ to reduce the communication frequency and lower the communication loss, e.g., adding a ${1}/{C}$ multiplicative factor to the listed choice of $f(p)$.}
 
\section{Proofs for Regret Analysis}
\subsection{Proof of Lemma \ref{lem:tp_event}}
\begin{proof}
	{To decouple the randomness of $A_m(p)$, we assume a virtual system without elimination, i.e., in this virtual system $\forall m\in[M], \forall p, A_m(p)=[K]$.} At phase $p$, $\forall m\in [M], \forall k\in A_m(p)$, $\bar{\mu}'_{k,m}(p)$ can be decomposed as
	\begin{equation*}
	\begin{aligned}
		\bar{\mu}'_{k,m}(p) = \left(\alpha+\frac{1-\alpha}{M}\right)\bar{\mu}_{k,m}(p)+\frac{1-\alpha}{M}\sum_{n\neq m}\bar{\mu}_{k,n}(p).
	\end{aligned}
	\end{equation*}
	It can be  shown that $\bar{\mu}_{k,m}(p)$ is a $\sqrt{\frac{1}{{N_{k,m}(p)}}}$-subgaussian random variable, since client $m$ has explored arm $k$ for $N_{k,m}(p)=\sum_{q=1}^p n_{k,m}(q)$ times in the global and local exploration sub-phases. However, $\forall n\in [M], n\neq m$, client $m$ can only make sure that $\bar{\mu}_{k,n}(p)$ is a $\sqrt{\frac{1}{N^g_{k,n}(p)}}$-subgaussian random variable, where $N^g_{k,n}(p)=\sum_{q=1}^pn^g_{k,n}(q)$, since she is only assured that each other client has explored arm $k$ in the global exploration sub-phases. Overall, we can claim that $\bar{\mu}'_{k,m}(p)$ is a $\sigma'_{k,m}(p)$-subgaussian random variable where 
	\begin{align*}
		\sigma'_{k,m}(p) &= \sqrt{\left(\alpha+\frac{1-\alpha}{M}\right)^2\frac{1}{N_{k,m}(p)}+\left(\frac{1-\alpha}{M}\right)^2\sum_{n\neq m}\frac{1}{N^g_{k,n}(p)}}\\
		&\leq \sqrt{\left(\alpha+\frac{1-\alpha}{M}\right)^2\frac{1}{[(1-\alpha)+M\alpha]F(p)}+\left(\frac{1-\alpha}{M}\right)^2\sum_{n\neq m}\frac{1}{(1-\alpha)F(p)}}\\
		& = \sqrt{\frac{1}{MF(p)}}.
	\end{align*}
	
	With the concentration inequality for subgaussian random variables, we have
	\begin{equation*}
		\Pb\left(|\bar{\mu}'_{k,m}(p)-\mu'_{k,m}|\geq B_{p}\right)\leq 2\exp\left\{-\frac{B^2_{p}}{2(\sigma'_{k,m}(p))^2}\right\}\leq 2\exp\left\{-\frac{\frac{4\log(T)}{MF(p)}}{2\frac{1}{MF(p)}}\right\} = \frac{2}{T^2}.
	\end{equation*}

	Thus, with the union bound, $P_{G}$ can be bounded as
	\begin{equation*}
	\begin{aligned}
		P_{G} &= 1-\Pb\left\{\exists p, \exists m\in [M], \exists k\in A_m(p), |\bar{\mu}'_{k,m}(p)-\mu'_{k,m}|\geq B_p\right\}\\
		&\geq 1-\sum_{p=1}^T\sum_{m=1}^M\sum_{k=1}^K\Pb\left(|\bar{\mu}'_{k,m}(p)-\mu'_{k,m}|\geq B_{p}\right)\\
		&\geq 1-\frac{2MK}{T}.
	\end{aligned}
	\end{equation*}	
	{Since this argument applies to $k\in [K]$, it also applies to all arms in the local active arm set $A_m(p)$ of the real system, which concludes the proof.}
\end{proof}

\subsection{Proof of Lemma \ref{lem:num_pull}}
\begin{proof}
	Recall that {$\forall k\neq k'_{*,m}$}, $p'_{k,m}$ is the smallest integer such that
	\begin{equation*}
		MF(p'_{k,m})\geq \frac{64\log(T)}{(\Delta'_{k,m})^2},
	\end{equation*}
	which ensures that $\forall p\geq p'_{k,m}, B_{p}\leq \frac{\Delta'_{k,m}}{4}$. Thus, based on that event ${G}$ happens, at phase $p'_{k,m}$, we have
	\begin{equation*}
	\begin{aligned}
		\bar{\mu}'_{k,m}(p'_{k,m})+B_{p'_{k,m}}&\overset{(i)}{\leq} \mu'_{k,m}+2B_{p'_{k,m}} \leq  \mu'_{k,m}+\frac{\Delta'_{k,m}}{2}\\
		& = \mu'_{*,m} -\frac{\Delta'_{k,m}}{2}\overset{(ii)}{\leq}  \bar{\mu}'_{k'_{*,m},m}(p'_{k'_{*,m},m})+B_{p'_{k,m}}-\frac{\Delta'_{k,m}}{2}\leq \bar{\mu}'_{k'_{*,m},m}(p'_{k'_{*,m},m})-B_{p'_{k,m}},
	\end{aligned}
	\end{equation*}
	where inequalities (i) and (ii) are guaranteed by event ${G}$. Thus, arm $k$ is guaranteed to be eliminated at phase $p'_{k,m}$ by client $m$.
\end{proof}

\subsection{Proof of Lemma \ref{lem:regret_expr}}
\begin{proof}
	Lemma \ref{lem:num_pull} indicates for a sub-optimal arm $k$, after phase $p'_{k,m}$, it is guaranteed to be eliminated from set $A_m(p)$. Thus, it is pulled for at most $\sum_{p=1}^{p'_{k,m}}\lceil \alpha M f(p)\rceil$ times in the local exploration sub-phases, which leads to the local exploration loss as
	\begin{equation*}
		R^{expr}_{l}(T)\leq \sum_{m=1}^M\sum_{k\neq k'_{*,m}}\Delta'_{k,m}\sum_{p=1}^{p'_{k,m}}\lceil \alpha M f(p) \rceil.
	\end{equation*}

	However, arm $k$ is still pulled in the global exploration sub-phases until $k\notin A(p)$, i.e., arm $k$ is eliminated by all of the clients whose optimal arm is not it. Since arm $k$ is guaranteed to be eliminated globally by phase $p'_{k}=\max_{m\in[M]}\{p'_{k,m}\}$, it is pulled for at most $\sum_{p=1}^{p'_{k}}\lceil (1-\alpha) f(p) \rceil$ times in the global exploration sub-phases. Thus, the global exploration loss can be bounded as:
	\begin{equation*}
		R^{expr}_{g}(T)\leq \sum_{m=1}^M\sum_{k\neq k'_{*,m}}\Delta'_{k,m}\sum_{p=1}^{p'_{k}}\lceil (1-\alpha) f(p) \rceil.
	\end{equation*}
\end{proof}

\subsection{Proof of Lemma \ref{lem:regret_expt}}
\begin{proof}
	At phase $p$, the exploitation time for client $m$ is at most $\max_n\{|A_n(p)|-A_m(p)\}\left\lceil M\alpha f(p)\right\rceil$, which is the difference between the longest local exploration duration and her local exploration duration. The probability that the exploited arm in the exploitation phase, i.e., arm $\bar{k}'_{*,m}$, is arm $k$ instead of $k'_{*,m}$ can be bounded as:
	\begin{align*}
		 \Pb\left(\bar{k}'_{*,m}=k\right) &\leq P\left(\bar{\mu}'_{k'_{*,m},m}({p-1})\leq \bar{\mu}_{k,m}({p-1})\right)\\
		& = P\left(\bar{\mu}'_{k'_{*,m},m}({p-1})-\bar{\mu}_{k,m}({p-1})-\Delta'_{k,m}\leq -\Delta'_{k,m}\right)\\
		& \overset{(i)}{\leq} 2\exp\left\{-\frac{(\Delta'_{k,m})^2}{2(\sigma^{'2}_{k,m}({p-1})+\sigma^{'2}_{k'_{*,m},m}({p-1}))}\right\}\\
		&\leq 2\exp\left\{-\frac{(\Delta'_{k,m})^2MF(p-1)}{4}\right\}\\
		&=P'_{k,m}(p).
	\end{align*}
	
	Thus, it can be shown that the exploration loss caused by arm $k$ for client $m$ is bounded as
	\begin{align*}
		R^{expt}_{k,m}(T)&\leq \Delta'_{k,m}\sum_{p=1}^{p'_{k,m}}\left(\max_n\{|A_n(p)|-A_m(p)\}\right)\left\lceil M\alpha f(p)\right\rceil P'_{k,m}(p) \\
		&\leq  \Delta'_{k,m}\sum_{p=1}^{p'_{k,m}}K\left\lceil M\alpha f(p)\right\rceil\exp\left\{-\frac{(\Delta'_{k,m})^2MF(p-1)}{4}\right\}.
	\end{align*}
	The overall exploration loss can be obtained by summing over all of the clients and arms:
	\begin{equation*}
		R^{expt}(T) = \sum_{m=1}^M\sum_{k=1}^K\Delta'_{k,m}R^{expt}_{k,m}(T)\leq \sum_{m=1}^M\sum_{k\neq k'_{*,m}}\sum_{p=1}^{p'_{k,m}}K\left\lceil M\alpha f(p)\right\rceil\Delta'_{k,m}\exp\left\{-\frac{(\Delta'_{k,m})^2MF(p-1)}{4}\right\}.
	\end{equation*}
	
	{In addition, we note that in phase $p=1$, all the players share the same global and local active arm sets, i.e., $\forall m\in [M], A_m(p)=A(p) = [K]$, which means there would be no exploration loss. Thus, the sum of index $p$ in the exploitation loss above can start from $2$ instead of $1$. This fact does not change the scaling of the overall regret, but would be useful in deriving Corollary~\ref{col:regret_overall_log} from Theorem~\ref{thm:regret_overall}.}
\end{proof}

\subsection{Proof of Lemma \ref{lem:regret_comm}}
\begin{proof}
	As designed in the \textsc{PF-UCB} algorithm, clients do not communicate any more after they find their optimal arms. Thus, there is no more communication after phase $p'_{\max}=\max_{k\in[K]}\{p'_{k,m}\} = \max_{m\in [M]}\max_{k\neq k'_{*,m}}\{p'_{k,m}\}$. Before phase $p'_{\max}$, there are two communications in each phase for arm statistics and active sets, respectively, which leads to the communication loss upper bound as:
	\begin{equation*}
		R^{comm}(T)\leq 2CMp'_{\max}.
	\end{equation*}
\end{proof}

\subsection{Proof of Theorem \ref{thm:regret_overall}}
\begin{proof}
	Lemmas \ref{lem:regret_expr}, \ref{lem:regret_expt} and \ref{lem:regret_comm} are all based on the condition that event ${G}$ happens, which has probability $P_{G}$ as shown in Lemma \ref{lem:tp_event}. When event ${G}$ does not happen, the regret is directly upper bounded by $MT+2CMT$, which assumes full exploration and communication loss. Thus, Theorem \ref{thm:regret_overall} follows by putting everything together as:
	\begin{align*}
		& R(T) = P_{G} \left(R^{expr}(T)+R^{expt}(T)+R^{comm}(T)\right)+(1-P_{G})(1+2C)MT\\
		&\leq R^{expr}_l(T)+R^{expr}_g(T)+R^{expt}(T)+R^{comm}(T)+2M^2K(1+2C)\\
		&\leq \sum_{m=1}^M\sum_{k\neq k'_{*,m}}\Delta'_{k,m}\sum_{p=1}^{p'_{k,m}}\lceil \alpha M f(p) \rceil+\sum_{m=1}^M\sum_{k\neq k'_{*,m}}\Delta'_{k,m}\sum_{p=1}^{p'_{k}}\lceil (1-\alpha) f(p) \rceil\\
		&+\sum_{m=1}^M\sum_{k\neq k'_{*,m}}\Delta'_{k,m}\sum_{p=1}^{p'_{k,m}}K\left\lceil M\alpha f(p)\right\rceil\exp\left\{-\frac{(\Delta'_{k,m})^2MF(p-1)}{4}\right\}+2CMp'_{\max}+2M^2K(1+2C).
	\end{align*}
\end{proof}

\subsection{Proof of Corollary \ref{col:regret_overall_log}}
\begin{proof}
	With $f(p)=2^p\log(T)$, $p'_{k,m}$ can be bounded from Eqn.~\eqref{eqn:num_pull} as
	\begin{equation*}
	p'_{k,m} = O\left(\log_2\left(\frac{64}{M(\Delta'_{k,m})^2}\right)\right).
	\end{equation*}
	Plugging this into Theorem \ref{thm:regret_overall}, Corollary \ref{col:regret_overall_log} follows.
\end{proof}

\begin{figure}
	\centering
	\includegraphics[width=0.5\linewidth]{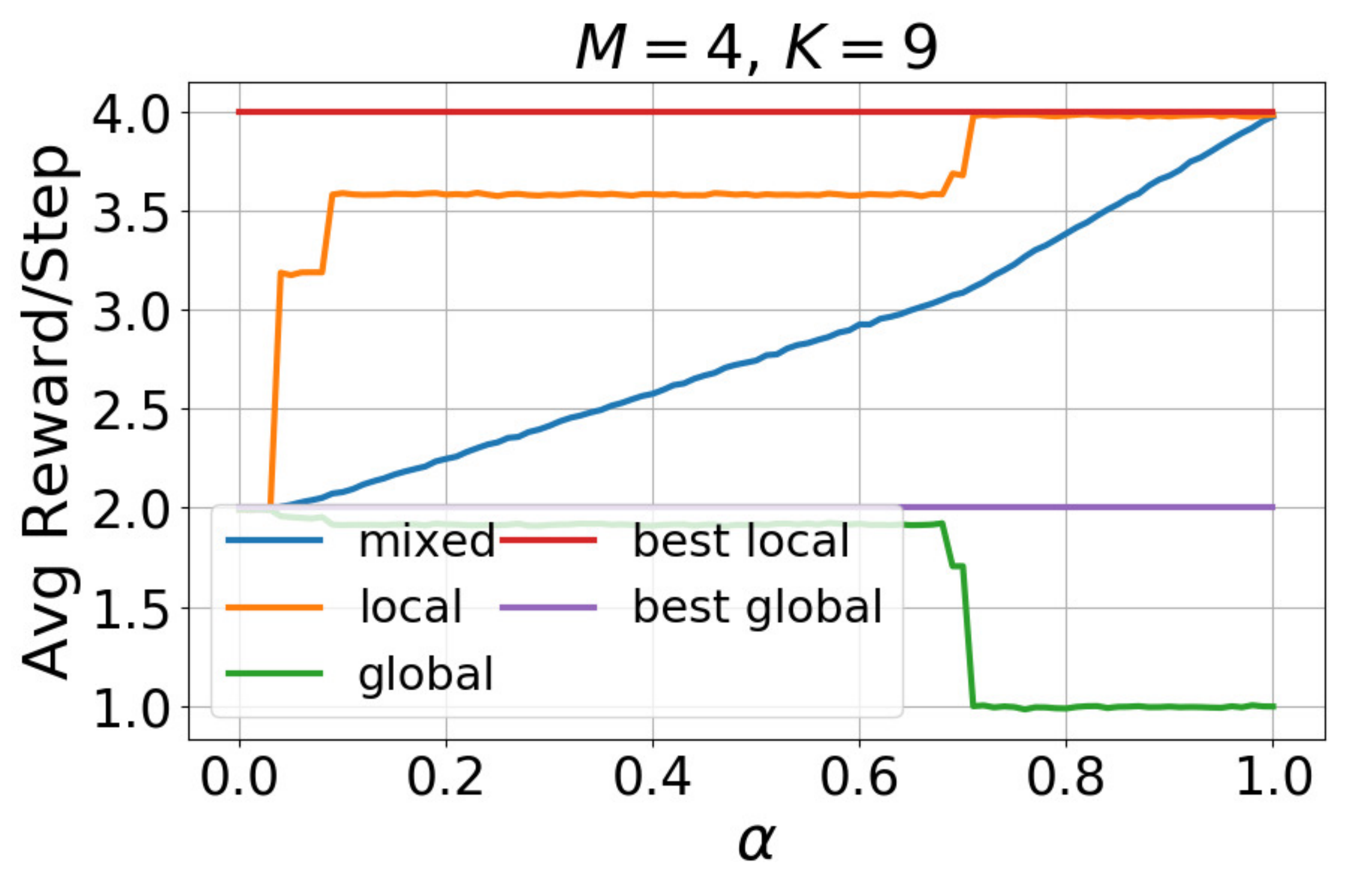}
	\caption{Synthetic Reward}
	\label{fig:syn_reward}
\end{figure}

\begin{figure}
	\centering
	\includegraphics[width=0.5\linewidth]{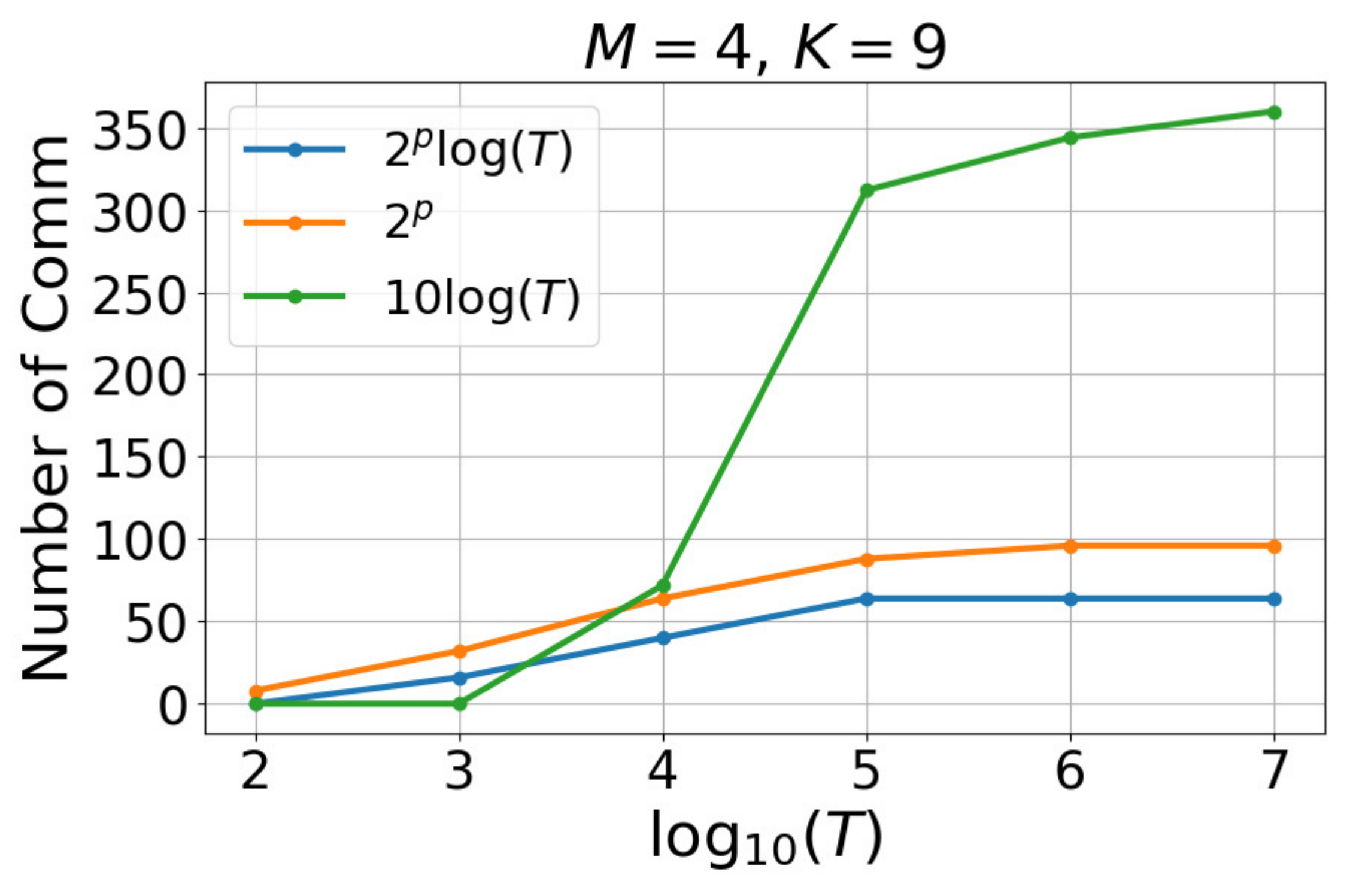}
	\caption{Number of Communications}
	\label{fig:syn_comm}
\end{figure}

\begin{figure}
	\centering
	\includegraphics[width=0.5\linewidth]{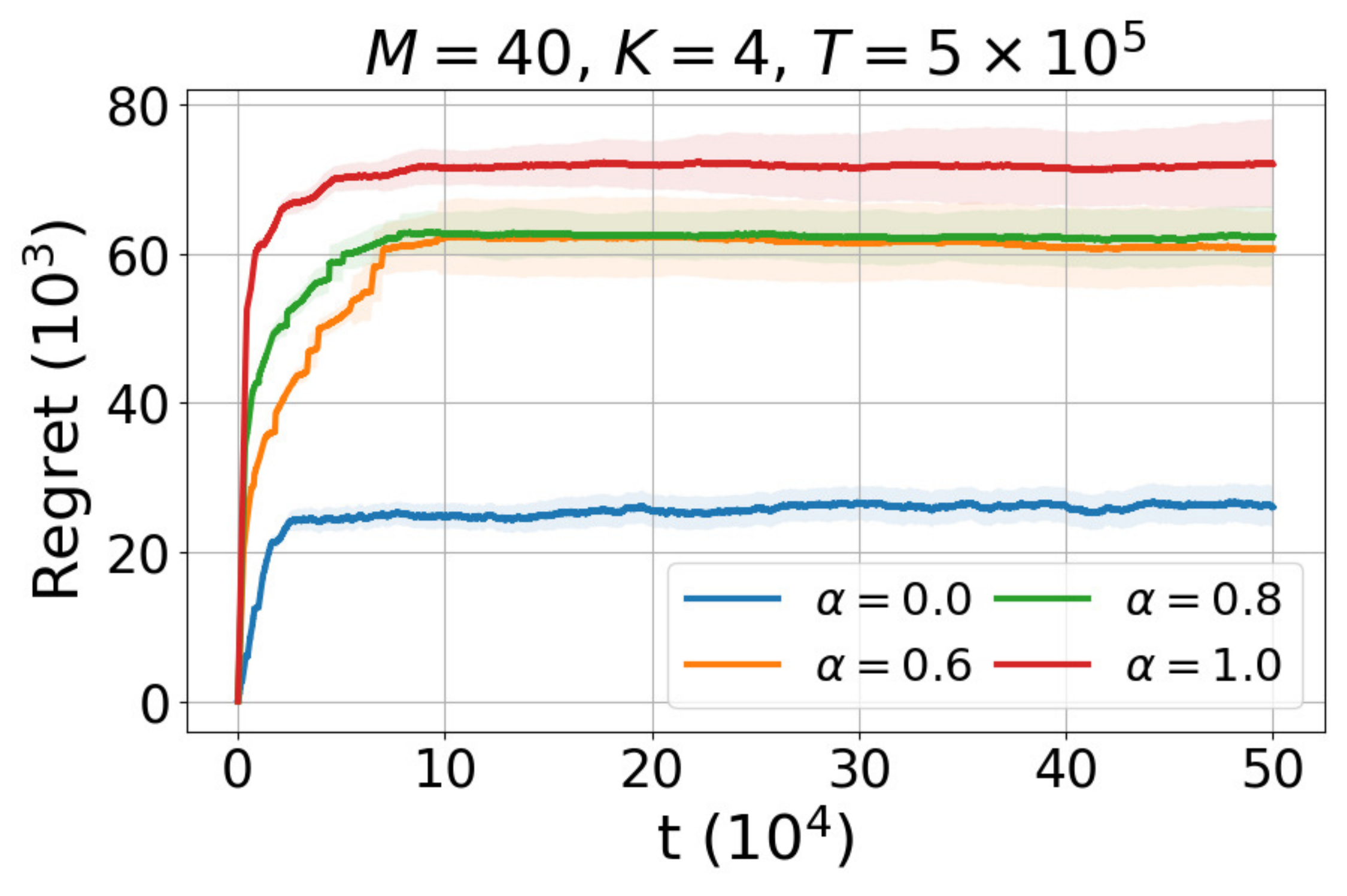}
	\caption{Large $M$ and Small $K$}
	\label{fig:large_m}
\end{figure}

\section{Additional Experimental Results}
The implementation codes of the PF-UCB and its enhancement used for simulations have been made publicly available at \url{https://github.com/ShenGroup/PF_MAB}, which also contains the synthetic dataset and the pre-processed real-world MovieLens dataset. 
The original version of the MovieLens dataset is publicly available at \url{https://grouplens.org/datasets/hetrec-2011/}. 

Experimental details and additional experiment results are provided here. First, for the synthetic dataset used in Fig.~\ref{fig:syn_regret}, the specific arm statistics are given as follows:
\begin{align*}
   \begin{bmatrix}
     1 & 0& 0 & 0 & 0.9 & 0.4 & 0.35 & 0.35 & 0.5\\
                  0& 1& 0& 0& 0.3& 0.9& 0.35& 0.3& 0.5\\
                  0& 0& 1& 0& 0.35& 0.35& 0.9& 0.3& 0.5\\
                  0& 0& 0& 1& 0.4& 0.3& 0.35& 0.9& 0.5
\end{bmatrix}
\end{align*}
where the rows and columns correspond to the clients and arms, respectively. This dataset is specially designed so that the local optimal arm for client $m\in\{1,2,3,4\}$ is arm $m$, while the global optimal arm is arm $9$. Moreover, each of the local optimal arms perform poorly at other clients. All remaining arms share similar global utilities, but diverge locally. The averaged per-step reward with PF-UCB under this synthetic dataset is reported in Fig.~\ref{fig:syn_reward}, which shows a similar trend as in Fig.~\ref{fig:movielens_reward}.

The communication {times in the horizon of $10^6$} for the synthetic game are provided in Table \ref{tbl:syn_comm_loss}. Compared with {the time horizon, the overall communication times are almost negligible}, which shows the efficiency of communication under the choice of $f(p)=2^p\log(T)$. The communication times under different time horizons for different choices of $f(p)$ are reported in Fig.~\ref{fig:syn_comm} with the same synthetic game and $\alpha=0.5$, which illustrates that $f(p)=10\log(T)$ leads to {more communications} for large $T$ {than the other two choices} and $f(p)=2^p\log(T)$ {is the most efficient one}. This observation coincides with the results in Table \ref{tbl:regret_P_Fed_UCB}.

\begin{table}[htb]	
    \centering
    \begin{tabular}{cc}  
      $\alpha$ & Comm Times \\ \hline
      $0$    & $104$ \\ 
      $0.2$   & $64$ \\ 
      $0.5$ & $72$ \\ 
      $0.9$  & $80$ \\ 
      $1$ & $56$\\
      \hline 
    \end{tabular} 
    \caption{Synthetic Communication Times} 
    \label{tbl:syn_comm_loss} 
\end{table}%
  
As in real-world FL systems, it is common to have a small $K$ (number of arms) and a large $M$ (number of clients). Additional experiments are performed with a small $K=4$ and a large $M=40$ with results reported in Fig.~\ref{fig:large_m}. It can be observed that PF-UCB still achieves stable performance in this scenario. 

\bibliography{MAB,FED,Shen}

\end{document}